%% file: main.tex
\documentclass[10pt,twocolumn,letterpaper]{article}

\usepackage{cvpr}              % To produce the CAMERA-READY version
\usepackage{multirow}
\input{preamble}

\definecolor{cvprblue}{rgb}{0.21,0.49,0.74}
\usepackage[pagebackref,breaklinks,colorlinks,citecolor=cvprblue]{hyperref}

\def\paperID{6697} % *** Enter the Paper ID here
\def\confName{CVPR}
\def\confYear{2024}

\title{SNI-SLAM: Semantic Neural Implicit SLAM}

\author{
  Siting Zhu\textsuperscript{1}\thanks{Equal Contribution.}, Guangming Wang\textsuperscript{2}\footnotemark[1], Hermann Blum\textsuperscript{3}, Jiuming Liu\textsuperscript{1},\\
  Liang Song\textsuperscript{4}, Marc Pollefeys\textsuperscript{3},
  Hesheng Wang\textsuperscript{1}\thanks{Corresponding Author.}\\
  {\textsuperscript{\rm 1} Department of Automation, Shanghai Jiao Tong University}
  {\textsuperscript{\rm 2} University of Cambridge}\\
  {\textsuperscript{\rm 3} ETH Zürich}
  {\textsuperscript{\rm 4} China University of Mining and Technology, China} \\
  {\tt\small \{zhusiting,liujiuming,wanghesheng\}@sjtu.edu.cn} \; {\tt\small gw462@cam.ac.uk} \\
  {\tt\small \{hermann.blum,marc.pollefeys\}@inf.ethz.ch} \; {\tt\small TS21060167P31@cumt.edu.cn}
}

\begin{document}
% \maketitle
\definecolor{myyellow}{RGB}{255,140,0}
\definecolor{myred}{RGB}{192,0,0}
\definecolor{myblue}{RGB}{30,144,255}
\twocolumn[{%
    \renewcommand\twocolumn[1][]{#1}%
    \setlength{\tabcolsep}{0.0mm} %0
    \maketitle
    \begin{center}
        \newcommand{\teaserwidth}{\textwidth}
    \vspace{-0.4in}
        \includegraphics[width=0.98\linewidth]{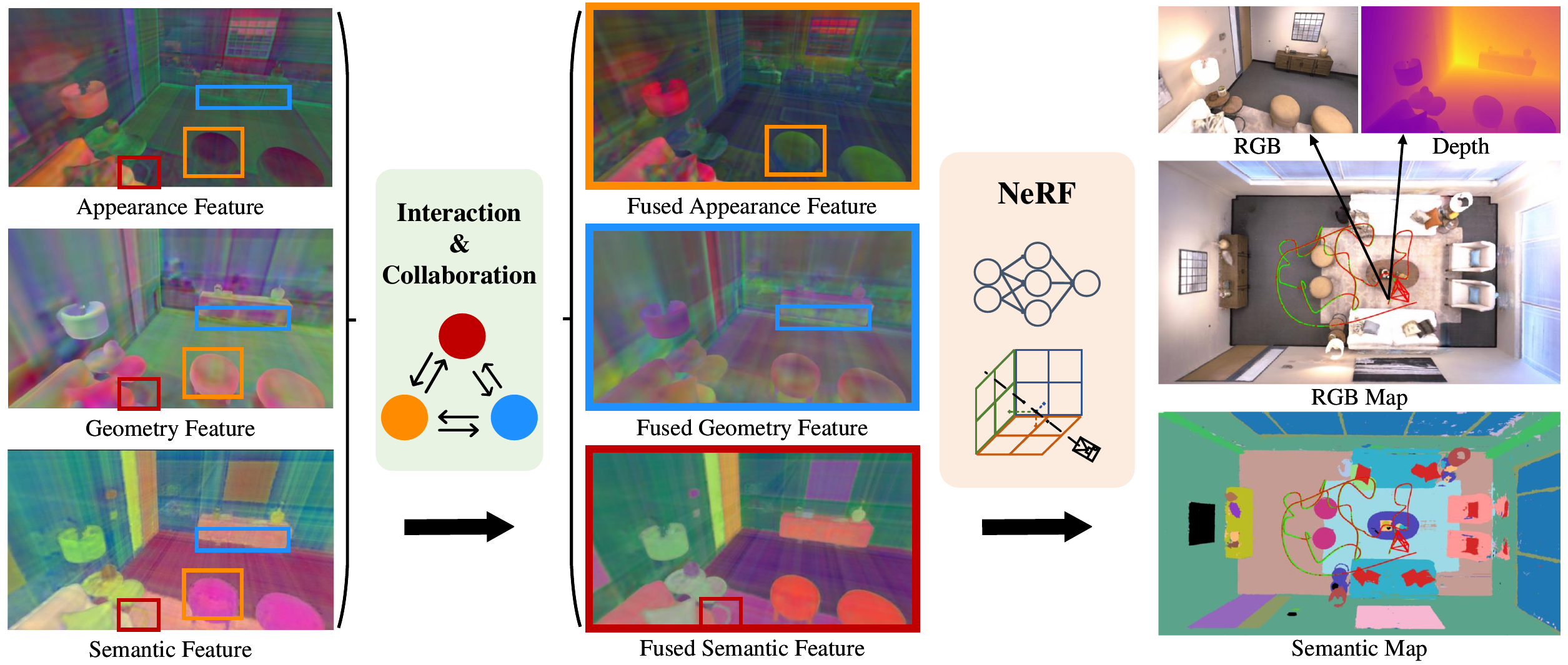}
      \vspace{-0.05in}
        \captionof{figure}{
        Our SNI-SLAM leverages the correlation of multi-modal features in the environment to conduct semantic SLAM based on Neural Radiance Fields (NeRF). This modeling strategy achieves not only higher accuracy compared with existing NeRF-based SLAM, but also enables real-time semantic mapping. We propose a feature collaboration method between appearance, geometry, and semantics, which significantly enhances the feature representation capabilities.
        \textcolor{myyellow}{\textbf{Fused Appearance (orange box):}} Shadowing on the chair caused by light is eliminated. 
        \textcolor{myblue}{\textbf{Fused Geometry (blue box):}} The inconsistency of the cabinet bottom edge is improved.  
        \textcolor{myred}{\textbf{Fused Semantic (red box):}}  The distinction between table leg and floor is enhanced.
        }
    \label{fig:teaser}
    \end{center}
}]

\renewcommand{\thefootnote}{\fnsymbol{footnote}}
\footnotetext{\footnotemark[1]Equal Contribution. \footnotemark[2]Corresponding Author.}

\input{sec/0_abstract}

\input{sec/final}

{
    \small
    \bibliographystyle{ieeenat_fullname}
    \bibliography{main}
}


\end{document}

% \maketitle

\input{sec/X_suppl}

{
    \small
    \bibliographystyle{ieeenat_fullname}
    \bibliography{main}
}

%% file: preamble.tex
\usepackage[dvipsnames]{xcolor}

%% file: sec/0_abstract.tex
\begin{abstract}
We propose SNI-SLAM, a semantic SLAM system utilizing neural implicit representation, that simultaneously performs accurate semantic mapping, high-quality surface reconstruction, and robust camera tracking. In this system, we introduce hierarchical semantic representation to allow multi-level semantic comprehension for top-down structured semantic mapping of the scene. In addition, to fully utilize the correlation between multiple attributes of the environment, we integrate appearance, geometry and semantic features through cross-attention for feature collaboration. This strategy enables a more multifaceted understanding of the environment, thereby allowing SNI-SLAM to remain robust even when single attribute is defective. Then, we design an internal fusion-based decoder to obtain semantic, RGB, Truncated Signed Distance Field (TSDF) values from multi-level features for accurate decoding. Furthermore, we propose a feature loss to update the scene representation at the feature level. Compared with low-level losses such as RGB loss and depth loss, our feature loss is capable of guiding the network optimization on a higher-level. Our SNI-SLAM method demonstrates superior performance over all recent NeRF-based SLAM methods in terms of mapping and tracking accuracy on Replica and ScanNet datasets, while also showing excellent capabilities in accurate semantic segmentation and real-time semantic mapping. Codes will be available at \href{https://github.com/IRMVLab/SNI-SLAM}{https://github.com/IRMVLab/SNI-SLAM}.

% The effectiveness of our SNI-SLAM are demonstrated on Replica and ScanNet datasets. Our method outperforms all recent NeRF-based SLAM methods in mapping and tracking accuracy, while demonstrating an excellent semantic segmantation and semantic mapping accuracy. 
 
\end{abstract}

%% file: sec/final.tex
\vspace{-0.2in}
\section{Introduction}
\label{sec:intro}
\hspace*{12pt}
Dense semantic Simultaneous Localization and Mapping (SLAM) is a fundamental challenge in robotics~\cite{rosinol2020kimera, mccormac2017semanticfusion} and autonomous driving~\cite{tian2022kimera,liu2023regformer,wang2022efficient,wang2021hierarchical}. It incorporates semantic understanding of the environment into map construction and estimates camera pose simultaneously. Compared with traditional SLAM, semantic SLAM is capable of identifying, categorizing, and relating entities in the scene as well as generating semantic maps.  

Traditional semantic SLAM has limitations including its inability to predict unknown areas and high storage space requirements~\cite{liu2023unsupervised}. Recently,
Neural Radiance Fields (NeRF)~\cite{martin2021nerf} have shown remarkable capability in scene representation, promising to address these limitations. Compared with traditional SLAM mapping representations such as TSDF and point cloud, this implicit scene representation benefits from continuous modeling and low storage cost. Following the advantages of implicit representation, NeRF-based SLAM~\cite{imap,nice,yang2022vox,eslam,wang2023co} methods have been developed. However, most existing NeRF-SLAM systems establish RGB maps, where color information is not directly suitable for downstream tasks such as navigation. In the meantime, there has been some works~\cite{iLabel,zhi2021place} demonstrating that NeRF can jointly learn geometric and semantic representations. 
However, these works require hours of offline training to obtain semantic scene representation, which is impractical for semantic SLAM that inherently demands real-time performance. Therefore, developing a semantic SLAM system based on NeRF is essential and challenging.

For semantic NeRF-based SLAM, there are two challenges: 1) Appearance, geometry and semantic information are interrelated, so processing them independently will lose interact connections, leading to an incomplete understanding of the image or scene. 2)  As the appearance of a scene, such as color, varies under different views, leveraging semantic multi-view consistency to optimize appearance will affect the details of the appearance, and vice versa.  

For the first challenge, MSeg3D~\cite{li2023mseg3d} fuses geometry and semantic features to obtain more accurate semantic segmentation results. However, this work does not take advantage of appearance information as another modality to enhance semantic expression from the visual structural perspective. Moreover, mutual reinforcement of different modalities is not explored either. In this paper, we use the individual characteristics of appearance, semantics, and geometry, to design a mutual collaboration and enhancement approach between these modalities based on cross-attention. This design enables improvements for each modality respectively.

For the second challenge, Semantic-NeRF~\cite{zhi2021place} appends a segmentation renderer before injecting viewing directions into the Multi-layer Perceptron (MLP). However, the impact of semantic optimization on appearance and geometric expression is not explored. To address this challenge, we propose a one-way correlation approach between different modalities by improving the decoder design and rendering process. This allows valuable information from one modality to enhance other modalities without affecting the original representation or being influenced in the reverse.

Overall, we provide the following contributions:

\begin{itemize}
 \item We present SNI-SLAM, a dense RGB-D semantic SLAM system based on NeRF, which can achieve accurate 3D semantic segmentation by real-time mapping.
 We introduce hierarchical semantic encoding for precisely constructing semantic maps.
 In addition, we utilize a feature loss to guide the network optimization on a higher-level, resulting in superior scene optimization results.
 
\end{itemize}
\begin{itemize}
  \item We perform an advanced feature collaboration approach to integrate geometry, appearance, and semantic features based on cross-attention. This design enables mutual reinforcement between different features. Moreover, we introduce a new decoder for one-way correlation to achieve enhanced decoding results without mutual interference. 
\end{itemize}
\begin{itemize}
  \item Extensive evaluations are conducted on two challenging datasets, Replica~\cite{straub2019replica} and ScanNet~\cite{dai2017scannet}, to demonstrate our method attains state-of-the-art performance compared with existing NeRF-based SLAM in mapping, tracking, and semantic segmentation.
\end{itemize}

% More relevant to our work are vMAP~\cite{kong2023vmap} and Neural Implicit Dense Semantic SLAM~\cite{haghighi2023neural}. vMAP does not perform semantic mapping, and only utilizes semantic segmentation results for object association. Neural Implicit Dense Semantic SLAM performs semantic mapping, but it treats semantics simply as the same information as color, and no correlation is made between different features. Also, there is no evaluation on the accuracy of mesh reconstruction.
%-------------------------------------------------------------------------

\begin{figure*}
  \centering
  \includegraphics[width=\linewidth]{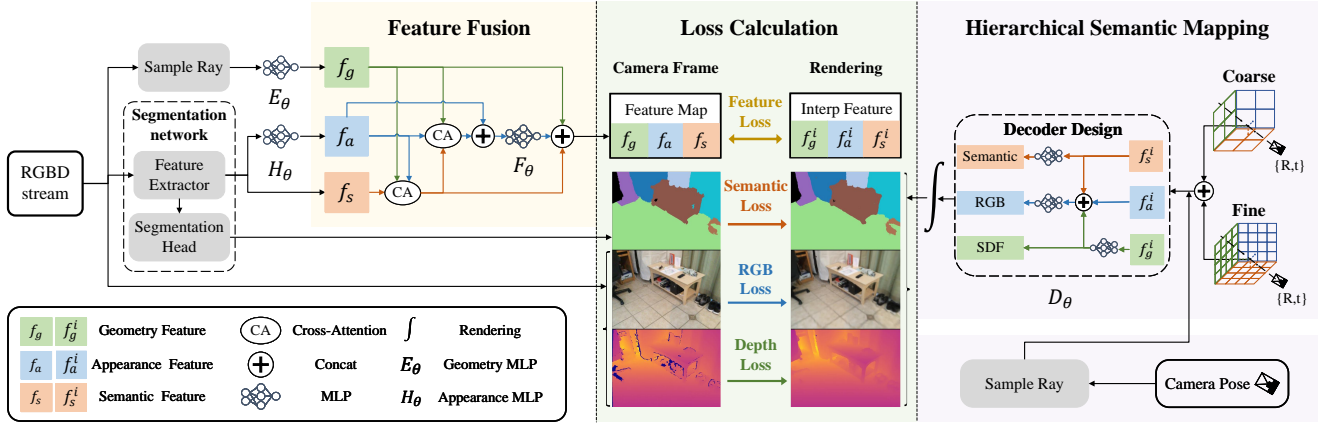}
  \caption{\textbf{An overview of SNI-SLAM.} Our method takes an RGB-D stream as input. RGB images are fed into semantic feature extractor to obtain semantic features. These features are then transformed into appearance features through appearance MLP \(H_{\theta}\). Geometry features are derived from ray sampling and then processed through geometry MLP \(E_{\theta}\). Subsequently, these three types of features are fused using cross-attention based feature fusion and generate feature map.
  This feature map, the input RGB-D, and the segmentation results obtained from segmentation network serve as supervision signals. Generated features are obtained by interpolation of scene representation, then these features are utilized for feature loss construction as well as to obtain the generated RGB, depth and semantics through decoding and rendering process. Supervision and generated information are used for loss construction to update scene representation and MLP network. 
  We use hierarchical semantic representation for semantic mapping. For camera tracking, we utilize loss functions to optimize camera pose. We follow \cite{eslam} for geometry and appearance scene representation.}
  \label{fig:overview}
  \vspace{-0.15in}
\end{figure*}

\section{Related Work}
\textbf{Semantic SLAM.}\hspace*{5pt}
Visual odometry~\cite{qin2018vins,ORB-SLAM3,davison2007monoslam,geneva2020openvins,mourikis2007multi, wang2020unsupervised} and real-time dense mapping~\cite{izadi2011kinectfusion,hornung2013octomap,whelan2015elasticfusion,dai2017bundlefusion,reijgwart2019voxgraph} are capable of localization and scene reconstruction. 
Semantic SLAM combines the advantages of visual odometry and real-time dense mapping, while integrating semantic information to achieve higher level understanding of the environment~\cite{mccormac2017semanticfusion}. This technology enables the robot to understand its own position and the meaning of the elements in the environment. SLAM++~\cite{salas2013slam++} is object-aware RGB-D SLAM that uses joint pose graph to represent object-level information in the scene. Kimera~\cite{rosinol2020kimera} relies on RGB-D or stereo sensing to generate dense semantic mesh maps and uses visual-inertial odometry for the motion estimation. 
These methods utilize explicit modeling for 3D semantic reconstruction. However, this representation requires considerable storage space and is insufficient for detailed reconstruction. In this paper, we leverage the advantages of neural implicit representaion for conducting high-fidelity semantic SLAM with minimal storage space.

\noindent\textbf{Neural implicit SLAM.}\hspace*{5pt}
Neural implicit representation~\cite{mildenhall2021nerf,oechsle2021unisurf,yariv2021volume,deng2023plgslam,deng2023prosgnerf} is a novel 3D representation approach that uses neural network to learn geometric representation and appearance information of the environment. This technique has a wide range of applications, such as new view synthesis~\cite{martin2021nerf,mildenhall2022nerf}, object pose estimation~\cite{wen2023bundlesdf,irshad2022shapo,huang2022neural,peng2022self,chen2023texpose} and surface reconstruction~\cite{wang2021neus,yariv2021volume,oechsle2021unisurf}. Neural implicit representation with SLAM is our main focus. 
iMAP~\cite{imap} introduces a single MLP network to achieve real-time mapping and localization of the scene. NICE-SLAM~\cite{nice} adopts hierarchical feature grid as scene representation, enabling more accurate mapping. ESLAM~\cite{eslam} uses multi-scale axis-aligned feature planes, reducing the memory consumption growth. Vox-Fusion~\cite{yang2022vox} is based on octree management for incremental mapping. Previous works have proved the feasibility for neural networks to model color and geometric information in the environment. However, the potential of neural implicit representation goes far beyond this, as it can be used to encode semantic information~\cite{zhi2021place, iLabel}. vMAP~\cite{kong2023vmap} is an object-level dense SLAM system that utilizes semantic segmentation results for object association, but it does not perform semantic mapping. 
NIDS-SLAM~\cite{haghighi2023neural} uses ORB-SLAM3~\cite{ORB-SLAM3} for tracking and Instant-NGP~\cite{mueller2022instant} for mapping. For the processing of semantic information, it maps the segmentation results to color encodings for optimization of network. However, this work does not integrate semantic with other features of the environment, such as geometry and appearance. In this paper, we introduce cross-attention based feature fusion to 
incorporate semantic, appearance, and geometry features, thus improving the accuracy of mapping, tracking, and semantic segmentaion.

%-------------------------------------------------------------------------
\section{Method}
\hspace*{12pt}The overview of our method is shown in Fig.~\ref{fig:overview}. Given an input RGB-D frames \(I = \{c_i, d_i\}_{i=1}^{N}\), we perform dense semantic mapping and real-time tracking by jointly optimizing the scene representation, the MLP network and camera pose.
Sec.~\ref{subsec:Feature Fusion} describes how to integrate geometric, semantic, and appearance features 
through feature fusion based on cross-attention. Sec.~\ref{subsec:Semantic Mapping} presents the hierarchical semantic mapping and localization process, including semantic representation, a new decoder design, volume rendering, and camera tracking. Sec.~\ref{subsec:Loss Functions} introduces the loss functions.

%-------------------------------------------------------------------------
\subsection{Cross-Attention based Feature Fusion}
\label{subsec:Feature Fusion} 
\hspace*{12pt}Geometry, semantics and appearance are interconnected. For semantic and appearance features, the appearance of an object may vary under changing light conditions or viewing angle, but its semantic feature usually remains the same. This stability makes semantic feature an important tool for recognizing and understanding objects. In the meantime, the appearance feature of an object can also enhance our understanding of its semantic information. By observing the color, brightness, or texture of an object, we can infer which category an object belongs to. For geometry and semantic, robots can recognize and use geometric feature to locate and quantify the position and shape of an object. This information can then be utilized to infer the likely nature or identity of the object. In addition, semantic information can be used to improve understanding of the geometry and location of objects.

Considering the correlation among features, we employ cross-attention to fuse geometry feature \(f_g\), semantic feature \(f_s\), and appearance feature \(f_a\).  The input RGB image is passed through a pretrained semantic segmentation network to obtain semantic feature \(f_s\). In this work, we utilize an universal feature extractor Dinov2~\cite{oquab2023dinov2}, followed by segmentation head to construct the segmentation network. The extracted semantic feature lacks specificity to the environment as it is derived from a pretrained segmentation network. Therefore, we utilize real-time updated appearance MLP \(H_{\theta}\) to transform the semantic feature into appearance feature \(f_g=H_{\theta}(f_s)\). This MLP network stores environment-specific appearance information.
For geometry feature, we first obtain the coordinates of 3D points \(\{p_i\}_{i=1}^{N}\) through ray sampling. Then, we use a NeRF-based frequency encoding~\cite{mildenhall2021nerf} to get vector \(\gamma(p)\):
\begin{flalign}
\scalebox{0.91}{
$\gamma(p)=(sin2^{\scriptscriptstyle 0} \pi p,cos2^{\scriptscriptstyle 0}\pi p, \ldots,sin2^{\scriptscriptstyle L-1} \pi p,cos2^{\scriptscriptstyle L-1} \pi p)$,}
\end{flalign}
where \(L\) defines the total count of frequencies used. We use \(L=6\) for 3D coordinates. \(\gamma(p)\) is processed through geometry MLP \(E_{\theta}(\gamma(p))\) to obtain geometry feature \(f_g\), which stores geometry information of the environment.

Then, we leverage the structural property of geometry to guide attention. \(f_g\) is used as \(Q\), \(f_a\) is used as \(K\), and \(f_s\) is used as \(V\), to perform cross-attention calculation to obtain fused semantic feature \(T_s\): 
\begin{equation}
T_s = softmax(\frac{f_gf_a^T}{\sqrt{||f_a||_2^2}})f_s .
\end{equation}
Through this fusion, the weighted combination of semantic information is dynamically adjusted based on geometry and appearance feature matches, thereby minimizing the influence of incorrect semantic predictions by highlighting matches and downplaying mismatches.
Moreover, we utilize \(f_a\), \(f_g\) and fused semantic features \(T_s\) as \(V\), \(Q\) and \(K\), to obtain fused appearance feature \(T_a\) respectively:
\begin{equation}
T_a = softmax(\frac{f_g \cdot T_s^T}{\sqrt{||T_s||_2^2}})f_a .
\end{equation}
The fused appearance feature \(T_a\) is enhanced through cross-attention, but it may lose some fine-grained details present in the original appearance feature \(f_a\). Therefore, we concatenate \(f_a\) and \(T_a\), and then pass concatenated feature through fusion MLP \(F_{\theta}\). This fusion preserves the augmented information from \(T_a\) while also integrating the finer details from \(f_a\), thus achieving a more enriched appearance representation.
Then, the result is concatenated with \(f_g\) and \(T_s\) to obtain feature map \(FM=\{f_g, T_a', T_s\}\).
This multi-modal feature fusion approach based on cross-attention facilitates interaction and mutual learning among features from different modalities, resulting in more accurate feature representation.
 
 %-------------------------------------------------------------------------
\begin{figure}
  \centering
  \includegraphics[width=\linewidth]{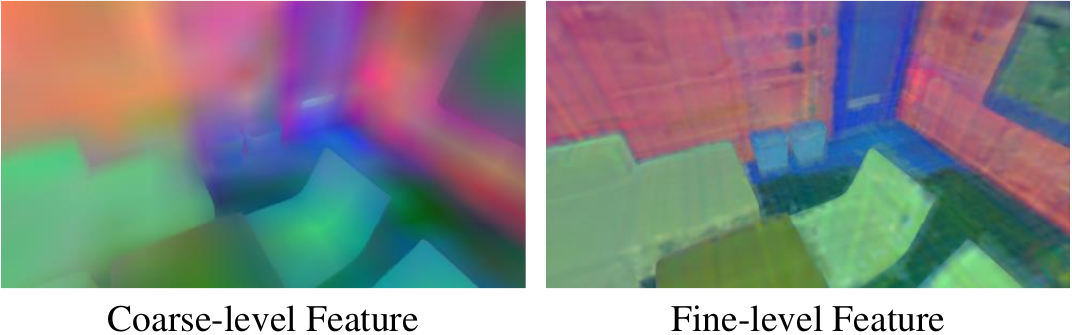}
  \caption{\textbf{Visualization of coarse-level and fine-level features.} Coarse-level feature captures general structure and arrangement of components. Fine-level feature provides more fine-grained details.}
  \label{fig:hsm}
  \vspace{-0.2in}
\end{figure}

\subsection{Hierarchical Semantic Mapping}
\label{subsec:Semantic Mapping} 
\hspace*{12pt} Currently, existing NeRF-based semantic modeling methods employ single-level neural implicit representation, regardless of whether they use voxel grid~\cite{vora2021nesf} or MLP~\cite{mazur2023feature,zhi2021place}. However, their performances are often limited when dealing with complex scenarios. We discover that using a hierarchical approach is more effective for semantic representation of the environment. When looking at a scene, we first grasp the overall layout and identify the main objects to develop a coarse understanding. After that, we shift our focus to more finely detailed. This top-down approach allows us to understand and process complex semantic information more naturally and efficiently.
Therefore, we employ coarse-to-fine semantic modeling for scene representation in this paper. Moreover, we design a fusion-based decoder to obtain semantic, color, SDF values, then achieve semantic, RGB, depth images through rendering process.

\noindent\textbf{Coarse-to-fine Semantic Representation.}\hspace*{5pt}
We utilize feature planes~\cite{eslam} to store features, which saves storage space compared with voxel grid~\cite{nice,yang2022vox}. For semantic mapping,
we employ a coarse-to-fine semantic representation. For each feature plane, we use two different levels of spatial resolution, where \(\{F_{s-xy}^{coarse}, F_{s-xz}^{coarse}, F_{s-yz}^{coarse}\}\) represent coarse level features, \(\{F_{s-xy}^{fine}, F_{s-xz}^{fine}, F_{s-yz}^{fine}\}\) represent fine level features, visualization of coarse and fine semantic features are shown in Fig.~\ref{fig:hsm}. For a given coordinate, we then concatenate the corresponding coarse and fine feature. We demonstrate empirically that the introduction of multi-level semantic representations improves the performance of  implicit semantic modeling and provides finer and richer semantic understanding.

\noindent\textbf{Decoder Design.}\hspace*{5pt}
There are typically two common designs for decoders in existing models. One approach~\cite{eslam}  uses separate decoder networks to process different features. Another approach~\cite{yang2022vox} utilizes the decoder network to obtain geometric and color information from a single feature. However, both approaches suffer because these decoders optimize independently without interaction.
In our work, we incorporate the idea of feature collaboration into decoder module to obtain SDF, RGB, and semantic values from  geometry, appearance and semantic features. Inside the decoder, we concatenate geometry feature with appearance and semantic features, then the concatenated feature passes through MLP network to obtain color decoding information. This design 
provides one-way correlation to ensure that improvement and application of the features occur only in one direction, thereby preventing mutual interference between the features. In addition, it also facilitates information exchange between features, improving the network's understanding of them. Considering the complexity of rich semantic categories, a larger hidden layer is necessary for comprehensive modeling. In this work, we use 256 dimension hidden layer for semantic decoding.

\noindent\textbf{Rendering.}\hspace*{5pt}  
We sample \(N\) points on the ray \(\{p_n\}_{i=1}^{N}\) to generate color \(c(p_n)\), semantic \(s(p_n)\) and TSDF \(d(p_n)\) values of these points through decoder \(D_{\theta}(p_n)\). Then, we use SDF-based rendering method proposed in StyleSDF~\cite{or2022stylesdf} to convert SDF values into volume densities:
\begin{equation}
\begin{aligned}
\sigma_g(p_n) = \frac{1}{\alpha_g} \cdot \text{Sigmoid} \left( -\frac{d(p_n)}{\alpha_g}\right) ,\\
\sigma_s(p_n) = \frac{1}{\alpha_s} \cdot \text{Sigmoid} \left( -\frac{d(p_n)}{\alpha_s}\right) ,
\end{aligned}
\end{equation}
where \(\alpha_g\) represents a learnable parameter that determines the level of sharpness along the surface boundary. Another learnable parameter \(\alpha_s\) is used for semantic rendering. Volume density \(\sigma_g(p_n)\) is subsequently utilized in rendering both the color and depth associated with each ray to obtain rendered color \(\hat{c}\) and depth \(\hat{d}\):
\begin{equation}
\begin{split}
w_{g} &= \exp \left( -\sum_{i=1}^{n-1} \sigma_g(p_i) \right) \left(1 - \exp(-\sigma_g(p_n))\right) ,\\
\hat{c} &= \sum_{n=1}^{N} w_{g} \cdot c(p_n) , \quad \hat{d} = \sum_{n=1}^{N} w_{g} \cdot z_n .
\end{split}
\end{equation}
In this context, \(z_n\) represents the depth of point \(p_n\) in relation to the camera's pose. \(\sigma_s(p_n)\) is used in semantic rendering and obtain rendered semantic  \(\hat{s}\):
\begin{equation}
\begin{gathered}
w_{s} = \exp \left( -\sum_{i=1}^{n-1} \sigma_s(p_i) \right) \left(1 - \exp(-\sigma_s(p_n))\right) ,\\
\hat{s} = \sum_{n=1}^{N} w_{s} \cdot s(p_n) .
\end{gathered}
\end{equation}

%-------------------------------------------------------------------------
\subsection{Loss Functions}
\label{subsec:Loss Functions} 
\hspace*{12pt} We sample \(M\) pixels from input images and refer to paper~\cite{azinovic2022neural} for the definition of free space loss. This loss compels the MLP network to predict values for points \(p \in P_{m}^{fs}\) that are positioned between the camera optical center and the truncation region of the surface:
\begin{equation}
 \mathcal{L}_{fs} = \frac{1}{|M|}\sum_{m\in M}\frac{1}{|P_{m}^{fs}|}\sum_{p \in P_{m}^{fs}}(d(p)-1)^2 .
\end{equation}
For points within the truncated region and close to the surface, we follow~\cite{eslam} for loss function:
\begin{equation}
 \mathcal{L}_{tr} = \frac{1}{|M|}\sum_{m\in M}\frac{1}{|P_{m}^{tr}|}\sum_{p \in P_{m}^{tr}}(z(p)+T\cdot d(p)-D(m))^2 ,
\end{equation}
where \(z(p)\) is the depth of point \(p\) on the plane in relation to the camera, \(T\) is truncation distance. \(D(m)\) is depth of the ray measured by the sensor. 
\(P_{m}^{tr}\) represents the set of points located within the truncation region on the ray \(m\).

\noindent\textbf{Semantic Loss.}\hspace*{5pt} 
For the supervision of semantic information, we use cross-entropy loss. It is worth noting that in the process of rendering semantic,  we detach the gradient to prevent the semantic loss from interfering with the optimization of geometry and appearance features:
\begin{equation}
 \mathcal{L}_{s} = -\sum_{m\in M}\sum_{l=1}^{L}p_{l}(m) \cdot \text{log} \hat{p_l}(m) ,
\end{equation}
where \(p_l\) represents multi-class semantic probability at class \(l\) of the ground truth map.

\noindent\textbf{Feature Loss.}\hspace*{5pt}
When only using color, depth, and semantic values as supervision signals, the MLP network will overly focus on less significant details and ignore some more salient features. To address this problem, feature loss is constructed and utilized to provide additional guidance for updating feature plane and MLP network. By providing direct supervision on intermediate features, this higher-level loss enables the scene representaion to preserve important details:
\begin{equation}
 \mathcal{L}_{f} = \| f_{\text{extract}} - f_{\text{interp}} \|_1 ,
\end{equation}
where \(f_{\text{extract}}\) represents the feature map generated in Sec. \ref{subsec:Feature Fusion}, \(f_{\text{interp}}\) stands for features obtained by the interpolation from the feature planes. The extracted features are more accurate and used as supervision signals.

\noindent\textbf{Color and Depth Loss.}\hspace*{5pt} 
The input is RGB-D frames containing ground truth RGB and depth values. We construct color and depth loss by comparing the rendered RGB and depth values with the ground truth values. These loss functions are then utilized for updating the network:
\begin{equation}
\begin{gathered}
\mathcal{L}_{c} = \frac{1}{|M|} \sum_{i=0}^{|M|}\|C_i-C_{i}^{gt}\| ,\\
\mathcal{L}_{d} = \frac{1}{|M|} \sum_{i=0}^{|M|}\|D_i-D_{i}^{gt}\| ,
\end{gathered}
\end{equation}
where \(C_i\), \(D_i\) are rendered RGB and depth values, \(C_{i}^{gt}\), \(D_{i}^{gt}\) are ground truth values.

The complete loss function \(\mathcal{L}\) is the weighted sum of the above losses:
\begin{equation}
\mathcal{L} = \lambda_{fs} \mathcal{L}_{fs}+\lambda_{tr} \mathcal{L}_{tr}+ \lambda_{s} \mathcal{L}_{s}+\lambda_{f} \mathcal{L}_{f}+\lambda_{c} \mathcal{L}_{c}+\lambda_{d} \mathcal{L}_{d},
\end{equation}
where \(\lambda_{fs}, \lambda_{tr}, \lambda_{s}, \lambda_{f}, \lambda_{c}, \lambda_{d}\) are weighting coefficients.

%-------------------------------------------------------------------------
\begin{table*}
    \centering
    \small
    \begin{tabular}{c|cccc|cc}
    \toprule
    Methods & \multicolumn{4}{c|}{Reconstruction} & \multicolumn{2}{c}{Localization}    \\
    & Depth L1[cm] $\downarrow$ & Acc.[cm] $\downarrow$ & Comp.[cm]  $\downarrow$ & Comp.Ratio(\%) $\uparrow$ & ATE Mean[cm] $\downarrow$ & ATE RMSE[cm] $\downarrow$ \\                      
    \midrule
    iMAP*~\cite{imap}  & 4.645 & 3.624 & 4.934 & 80.515 & 3.118 & 4.153 \\
    NICE-SLAM~\cite{nice} & 1.903 & 2.373 & 2.645 & 91.137 & 1.795 & 2.503 \\
    Vox-Fusion~\cite{yang2022vox} & 2.913 & \textbf{1.882} & 2.563 & 90.936 & 1.027 & 1.473 \\
    Co-SLAM~\cite{wang2023co} & 1.513 & 2.104 & 2.082 & 93.435 & 0.935 & 1.059\\
    ESLAM~\cite{eslam} & \underline{0.945} & 2.082 & \underline{1.754} & \underline{96.427} & \underline{0.545} & \underline{0.678} \\
    SNI-SLAM (Ours) & \textbf{0.766} & \underline{1.942} & \textbf{1.702} & \textbf{96.624} & \textbf{0.397} & \textbf{0.456}\\
    \bottomrule
    \end{tabular}
    %Quantitative comparison of our proposed SNI-SLAM with other existing NeRF-based visual dense SLAM methods on the Replica dataset~\cite{straub2019replica} for SLAM metrics
    \caption{Quantitative comparison of map reconstruction and localization accuracy for our proposed SNI-SLAM and other NeRF-based dense SLAM methods. The results are average of 8 scenes on the Replica dataset~\cite{straub2019replica}. To ensure more objectivity in the results, each scene is tested and averaged with five independent runs. Our work outperforms previous works, indicating that our semantic-SLAM system has promising SLAM performance. For the details of the evaluations for each scene, please refer to the supplementary.}
    \label{tab:replica_slam}
    \vspace{-0.05in}
\end{table*}
\begin{figure*}
  \centering
  \small
  \includegraphics[width=\linewidth]{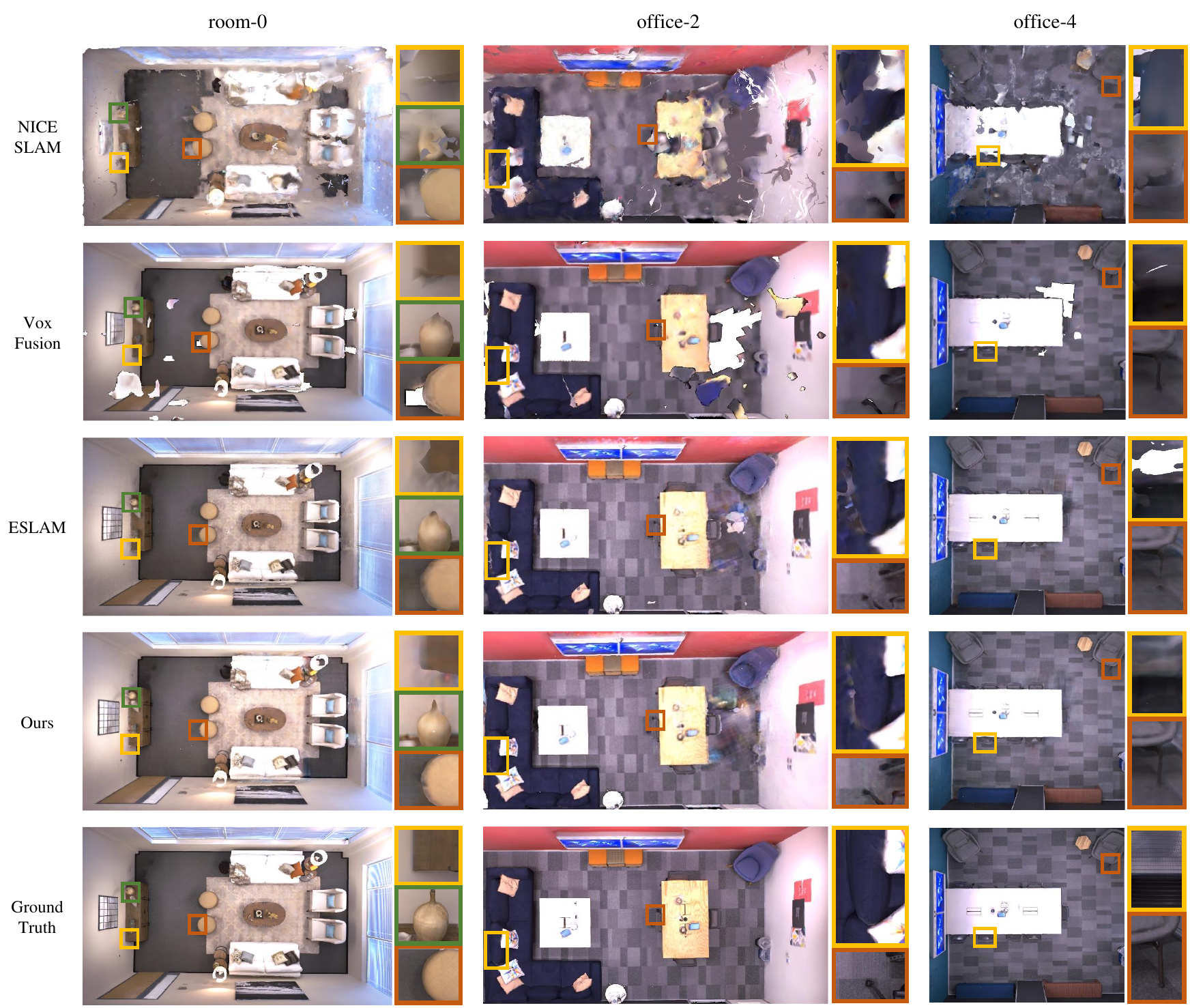}
  \caption{Qualitative comparison on scene reconstruction of our method and baseline. The ground truth images and details are rendered with ReplicaViewer software~\cite{straub2019replica}. We visualize 3 selected scenes of Replica dataset~\cite{straub2019replica} and details are highlighted with colorful boxes. Our method achieves more accurate detailed geometry and higher completion, especially in places that have limited observations.}
  \label{fig:detail_compare}
  \vspace{-0.1in}
\end{figure*}

\begin{table}
  \centering
  \footnotesize
  \begin{tabular}{l|ccccc}
    \toprule
    Methods & room0  & room1  & room2  & office0 & Avg. \\
     \midrule
     NIDS-SLAM~\cite{haghighi2023neural} & 82.45 & 84.08 & 76.99 & 85.94&82.37\\
     SNI-SLAM (Ours) &\textbf{88.42} & \textbf{87.43}& \textbf{86.16}& \textbf{87.63}&\textbf{87.41}\\
    \bottomrule
  \end{tabular}
   \vspace{-0.1in}
  \caption{
  Quantitative comparison of SNI-SLAM with existing semantic NeRF-based SLAM method NIDS-SLAM~\cite{haghighi2023neural} for semantic segmentation metrics \textit{mIoU(\%)} on 4 scenes of the Replica dataset~\cite{straub2019replica}, as its results are only reported on such scenes. For a fair comparison with NIDS-SLAM~\cite{haghighi2023neural}, the results are obtained using ground truth semantic label for supervision. For one scene, we calculate mIoU between rendered and ground truth semantic maps calculated every 50 frames to obtain average mIoU. Our method outperforms NIDS-SLAM~\cite{haghighi2023neural}. The performance of SNI-SLAM on other scenes is provided in the supplementary.
  }
  \vspace{-0.15in}
  \label{tab:replica_semantic}
\end{table}

\section{Experiments}
% \subsection{Experimental Setup}
\noindent\textbf{Datasets.}\hspace*{5pt}
We evaluate the performance of SNI-SLAM on three datasets, including 8 synthetic scenes on Replica~\cite{straub2019replica} and 4 real-world scenes on ScanNet~\cite{dai2017scannet}, both with semantic ground truth annotations, and 5 real-world scenes on TUM RGB-D~\cite{sturm2012benchmark} without semantic ground truth annotations.

\noindent\textbf{Metrics.}\hspace*{5pt}
To evaluate the SLAM system, we use metrics from Co-SLAM~\cite{wang2023co}. For mesh reconstruction metrics, we use \textit{Depth L1 (cm)}, \textit{Accuracy (cm)}, \textit{Completion (cm)}, and \textit{Completion ratio(\%)} with a threshold of 5cm. Also, we use ATE~\cite{sturm2012benchmark} for tracking accuracy evaluation. Semantic segmentation is evaluated with respect to mIoU~\cite{long2015fully} metric.

\noindent\textbf{Baselines.}\hspace*{5pt} 
We compare the metrics of the semantic segmentation accuracy with NIDS-SLAM~\cite{haghighi2023neural}, which is the only semantic NeRF-SLAM method to the best of our knowledge.
For SLAM accuracy, we compare our method with state-of-the-art NeRF-based dense visual SLAM methods~\cite{imap, nice, yang2022vox, eslam, wang2023co, sandstrom2023point}. For more detailed explanation, please refer to supplementary.

\noindent\textbf{Implementation Details.}\hspace*{5pt} 
% We adopt the coarse feature planes with a resolution of 24 cm and the fine feature planes with a resolution of 3 cm for semantic, geometry and appearance. We use 16-channel feature vectors to represent semantic, geometry and appearance features because we find that 16-channel vectors can already model the scene accurately. Increasing  feature channel can slightly improve accuracy, but it would lead to slower computation speed and require more storage space. 
We use 16-channel feature vectors to represent semantic, geometry and appearance features. The decoder MLP has two layers and the hidden layer dimension is 32. We run SNI-SLAM on NVIDIA RTX 4090 GPU. Please refer to the supplementary for further details of our implementation.

\subsection{Experimental Results}
\textbf{Replica dataset~\cite{straub2019replica}.}\hspace*{5pt}
 As shown in Tab.~\ref{tab:replica_slam}, our method achieves the highest accuracy compared with other NeRF-based SLAM methods and up to 32\% relative increase in tracking accuracy. \textit{Accuracy (cm)} is calculated based on the error between reconstructed points and ground truth points. Vox-Fusion~\cite{yang2022vox} achieves the highest \textit{Accuracy (cm)} because it only reconstructs observed areas and ignores errors in predicted unseen regions, but this strategy results in nearly worst \textit{Completion (cm)} and \textit{Completion ratio (\%)} metrics compared with other NeRF-SLAM methods.
 
 The reconstruction of 3 scenes are shown in Fig.~\ref{fig:detail_compare} with interesting regions highlighted with coloured boxes. For some narrow details, such as bottle necks and chair legs, other methods fail to correctly reconstrcut them and blend them into the background. Our method leverage semantic information to understand object categories, appearance cues to identify texture and materials, and geometric constraints to maintain valid shapes, thereby achieving complete modeling.
Moreover, other methods have difficulty reconstructing edges such as the corners of tables and the seams of sofas accurately. Our method incorporates three types of representations: appearance which has edge, color, texture information, geometry which has 3D structure information such as size, shape, position, and semantic representation which has advantages in distinguishing different object categories based on boundaries. Fusing these representations enables the network to model fine-grained details of objects, resulting in detailed reconstruction.  

As shown in Tab.~\ref{tab:replica_semantic}, our method outperforms NIDS-SLAM~\cite{haghighi2023neural} in segmentation metrics of all scenes and achieves up to 10\% increase on mIoU. 
% and top-view semantic mapping results are shown in Fig.~\ref{fig:semantic_color_vis}.

\noindent\textbf{ScanNet dataset~\cite{dai2017scannet}.}\hspace*{5pt} 
Following previous methods~\cite{nice,yang2022vox,wang2023co,eslam}, we evaluate tracking accuracy on the ScanNet dataset~\cite{dai2017scannet}. As shown in Tab.~\ref{tab:scannet}, our method also outperforms baseline methods and achieve 10\% improvement of accuracy in this real-world dataset. 
    
\begin{table}
  \centering
  \footnotesize
  \begin{tabular}{lccccc}
    \toprule
    Scene ID & 0000 & 0059 & 0106 & 0207 & Avg. \\
    \midrule
    iMAP*~\cite{imap}  & 55.95 & 32.06 & 17.50  & 11.91 & 29.36 \\
    NICE-SLAM~\cite{nice}  & 8.64 & 12.25 & 8.09 & 5.59 & 8.64 \\
    Co-SLAM~\cite{wang2023co} & 7.13 & 11.14 & 9.36 & 7.14 & 8.69 \\
    Vox-Fusion~\cite{yang2022vox} & 8.39 & 9.18 & 7.44 & 5.57 & 7.65  \\
    ESLAM~\cite{eslam} & 7.32 & 8.55 & 7.51 & 5.71 & 7.27 \\
    SNI-SLAM (Ours) & \textbf{6.90} & \textbf{7.38}  & \textbf{7.19} &  \textbf{4.70} & \textbf{6.54} \\
    \bottomrule
  \end{tabular}
  \vspace{-0.1in}
  \caption{We compare our proposed SNI-SLAM with other existing NeRF-based SLAM methods on ScanNet dataset~\cite{dai2017scannet} for tracking metrics \textit{RMSE (cm)}. Our method outperforms baseline.}
  \label{tab:scannet}
  \vspace{-0.1in}
\end{table}

\noindent\textbf{TUM RGBD dataset~\cite{sturm2012benchmark}.}\hspace*{5pt} 
As TUM dataset lacks semantic labels, we utilize SAM model DEVA~\cite{cheng2023tracking} for semantic segmentation to obtain 2D labels for semantic mapping and tracking. Compared with other NeRF-based SLAM methods, our method achieves up to 41\% improvement in Tab.~\ref{tab:tum}.

\begin{table}
  \centering
  \resizebox{\linewidth}{!}{
  \begin{tabular}{l|c c c c c c}
  \toprule
    \multirow{2}{*}{Method} & fr1/ & fr1/ & fr1/ & fr2/ & fr3/ & \multirow{2}{*}{Avg.}\\
    & desk & desk2 & room & xyz & office & \\
    % Method & fr1/desk & fr1/desk2 & fr1/room & fr2/xyz & fr3/office & Avg. \\
    \midrule
    NICE-SLAM~\cite{nice} & 4.26& 4.99 &34.49& 31.73& 3.87& 15.87\\
    Vox-Fusion~\cite{yang2022vox} & 3.52& 6.00& 19.53 &1.49 &26.01 &11.31\\
    Point-SLAM~\cite{sandstrom2023point} & 4.34 & 4.54 & 30.92 & 1.31 & 3.48 & 8.92\\
    SNI-SLAM (Ours) & \textbf{2.56} & \textbf{4.35} & \textbf{11.46} & \textbf{1.12} & \textbf{2.27}  & \textbf{4.35}\\
    \toprule
  \end{tabular}}
  \vspace{-0.12in}
   \caption{Comparison of our SNI-SLAM with other NeRF-based SLAM methods in tracking performance. We report \textit{RMSE (cm)} on 5 scenes of TUM RGBD dataset~\cite{sturm2012benchmark}.}
    \vspace{-0.1in}
  \label{tab:tum}
\end{table}

% \begin{figure}
%   \centering
%   \includegraphics[width=\linewidth]{images/semantic_color_vis.pdf}
%   \caption{Semantic reconstruction of 3 scenes on the Replica dataset~\cite{straub2019replica} are shown. Our method is capable of attaining relatively accurate results through optimization during the real-time mapping process, regardless of there existing some regions in the scene that have been little or no observed.  }
%   \label{fig:semantic_color_vis}
%   \vspace{-0.2in}
% \end{figure}

\begin{table}
  \centering
  \resizebox{\columnwidth}{!}{%
  \begin{tabular}{c|c|ccc|c}
    \toprule
    & Methods & Track. FPS~$\uparrow$ & Map. FPS~$\uparrow$& SLAM FPS~$\uparrow$ &\#param.~$\downarrow$\\
    \midrule
     \multirow{5}{*}{\rotatebox{90}{w/o sem}}& iMAP\cite{imap} & 9.92 & 2.23 & 1.822 & \textbf{0.26M} \\
    & NICE-SLAM~\cite{nice} & 13.70 & 0.20 & 0.198 & 12.2M \\
    & Vox-Fusion~\cite{yang2022vox} & 2.11 & 2.17 & 1.07 & 0.87M \\
    & Co-SLAM~\cite{wang2023co} & 17.24 & \textbf{10.20} & \textbf{6.41} & \textbf{0.26M} \\
    & ESLAM~\cite{eslam} & \textbf{18.11} & 3.62 & 3.02 & 6.85M \\
    \midrule
     \multirow{2}{*}{\rotatebox{90}{sem}}& NIDS-SLAM~\cite{haghighi2023neural} & --- & --- & 0.86 -- 2.13 & 12.6M \\
    & SNI-SLAM (Ours) & \textbf{16.03} & \textbf{2.48} & \textbf{2.15} & \textbf{6.2M} \\
    \bottomrule
  \end{tabular}%
  }
    \vspace{-0.1in}
  \caption{Runtime and memory comparison on Replica~\cite{straub2019replica} (w/o sem: without semantic mapping; sem: semantic mapping).  } 
  \label{tab:runtime}
  \vspace{-0.15in}
\end{table}

\subsection{Runtime Analysis}
\hspace*{12pt} We evaluate runtime and parameter numbers of our SNI-SLAM on Replica~\cite{straub2019replica} in Tab.~\ref{tab:runtime}. Our semantic NeRF-based SLAM is capable of semantic mapping with only a slight increase in runtime and similar parameter numbers compared with existing NeRF-based SLAM methods. Additionally, our method runs faster with half the number of parameters compared with exisitng baseline NIDS-SLAM~\cite{haghighi2023neural}.

\subsection{Ablation Study}
\hspace*{12pt}Tab.~\ref{tab:ablation_study} shows multiple experiments to validate the effectiveness of different component in SNI-SLAM.

\begin{table}
  \centering
  \footnotesize
  \begin{tabular}{l | cccc | cc}
    \toprule
    Name & HSM & FL & Dec & FF & RMSE[cm] & mIoU(\%)\\
    \midrule
    SR  &  &  &  &   & 0.83 & 71.5 \\
    HSR  & \checkmark &  &  &  & 0.55 & 84.1\\
    HSR+L & \checkmark & \checkmark &  &  & 0.47 & 85.0 \\
    DHSR+L & \checkmark & \checkmark & \checkmark &  & 0.43 & 85.3 \\
    SNI-SLAM & \checkmark & \checkmark & \checkmark & \checkmark & \textbf{0.33} & \textbf{86.0} \\
    \bottomrule
  \end{tabular}
   \vspace{-0.1in}
  \caption{Ablation study of our contributions on the office0 of Replica~\cite{straub2019replica} : (HSM) Hierarchical Semantic Mapping; (FL) Feature Loss; (Dec) Decoder Design; (FF) Cross-Attention based Feature Fusion; (SR) Semantic NeRF-based SLAM only with feature plane as scene representaion; (HSR) Add coarse-to-fine semantic mapping; (HSR+L), (DHSR+L) add corresponding innovation.}
  \label{tab:ablation_study}
  \vspace{-0.1in}
\end{table}

\begin{figure}
  \centering
  \includegraphics[width=\linewidth]{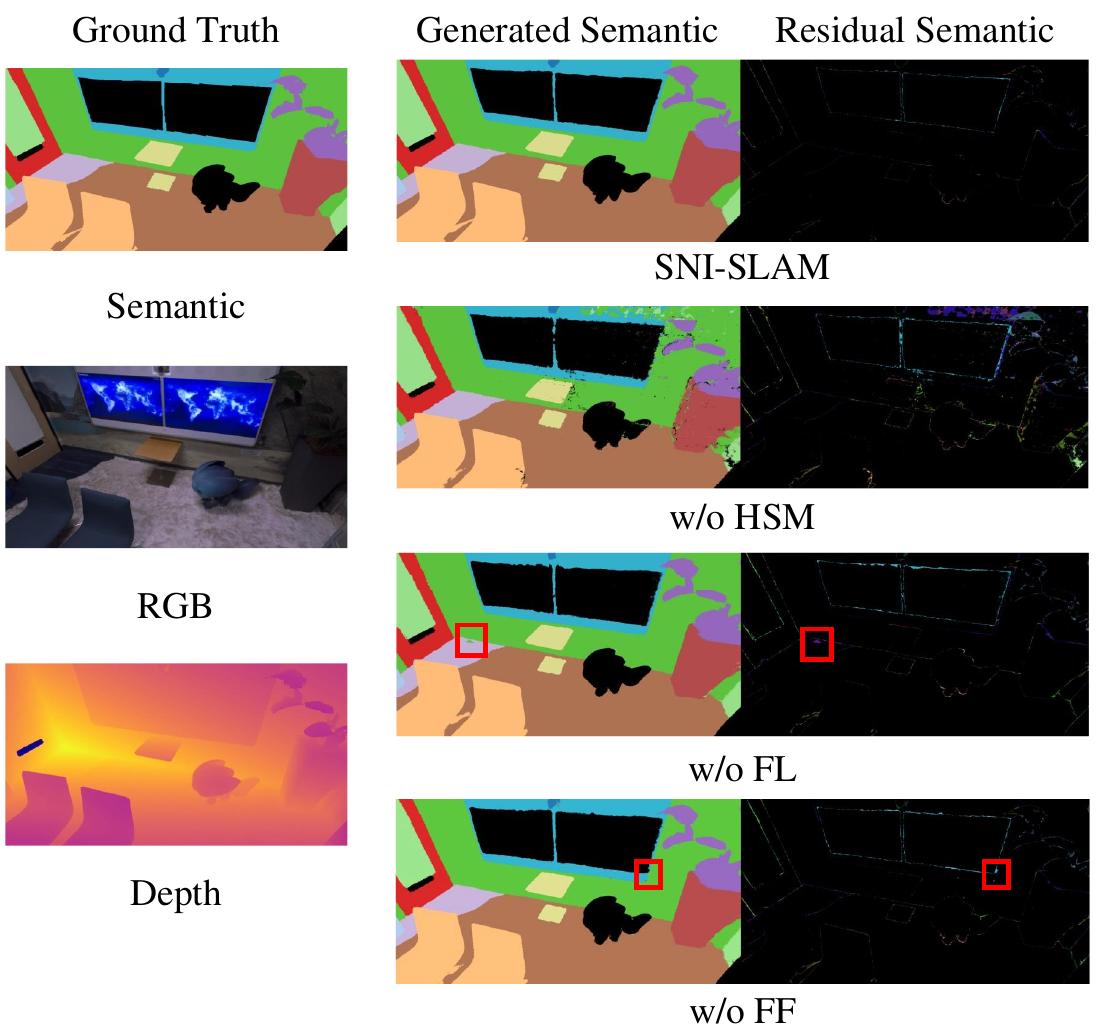}
  \vspace{-0.25in}
  \caption{Ablation study of semantic rendering results and ground truth labels on office0 of Replica~\cite{straub2019replica}. We visualize rendering results in different circumstances: (w/o HSM) without Hierarchical Semantic Mapping; (w/o FL) without Feature Loss; (w/o FF) without Feature Fusion. It can be seen from residuals that the whole SNI-SLAM achieves best semantic accuracy.}
  \label{fig:semantic_ablation_render}
  \vspace{-0.15in}
\end{figure}

\noindent\textbf{Hierarchical Semantic Mapping (HSM).} \hspace*{5pt}Tab.~\ref{tab:ablation_study} shows that coarse-to-fine semantic representation can significantly increase the accuracy of semantic mapping and tracking. 
Compared with single-layer representation, multi-layer semantic scene representation is capable of simultaneously taking into account the overall semantic and local semantic features. Fig.~\ref{fig:semantic_ablation_render} shows that single-layer representation may not adequately represent large semantic areas like walls. It could mis-segment walls into other labels by focusing too much on detailed information. In contrast, a hierarchical model can provide a more comprehensive understanding by representing both overall semantic categories and finer-grained details.
This semantic representation achieves more precise modeling and semantic expression.

\noindent\textbf{Feature Loss (FL).}
\hspace*{5pt} We validate the effectiveness of feature loss in Tab.~\ref{tab:ablation_study}. 
Constructing RGB, depth, semantic loss can only supervise information limited to one dimension, but features is capable of abstracting more information. Utilizing feature loss can force the model to learn important but easily ignored details in images, such as small objects or pixel details in edge regions. As shown in Fig.~\ref{fig:semantic_ablation_render}, semantic rendering results of whether to add feature loss reveals that constructing loss of geometry, appearance, and semantic features can avoid missegmentation at boundaries. 

% \noindent\textbf{Decoder Design (Dec).}
% \hspace*{5pt} Tab.~\ref{tab:ablation_study} demonstrates that fusing appearance and geometry information in decoder staegs enables the network to decode features more precisely.

\noindent\textbf{Cross-Attention based Feature Fusion (FF).}
\hspace*{5pt} The effectiveness of feature fusion module is validated in Tab.~\ref{tab:ablation_study}. Fig.~\ref{fig:semantic_ablation_render} displays that utilizing feature fusion can distinguish the TV screen from the background semantically. From RGB image, we can observe significant differences in color between the TV and its background, indicating a substantial divergence in their appearance features as well. Therefore, appearance feature can serve as a guidance to semantic feature through feature fusion, to avoid missegmentation in the cases where semantic segmentation network makes mistakes.
This fusion strategy leverages the complementarity between geometry, appearance, and semantic features, thereby generating a more powerful feature representation.

% \begin{table}
%   \centering
%   \scriptsize
%   \begin{tabular}{lccccccc}
%     \toprule
%     Scene ID & 0000 & 0059 & 0106 & 0169 & 0181 & 0207 & Avg. \\
%     \midrule
%     iMAP*  & 55.95 & 32.06 & 17.50 & 70.51 & 32.10 & 11.91 & 36.67 \\
%     NICE-SLAM & 8.64 & 12.25 & 8.09 & 10.28 & 12.93 & 5.59 & 9.63\\
%     Co-SLAM & 7.13 & 11.14 & 9.36 & 5.90 & 11.81 & 7.14 & 8.75 \\
%     Vox-Fusion & 8.39 & 9.18 & 7.44 & 6.53 & 12.20 & 5.57 & 8.22 \\
%     ESLAM & 7.32 & 8.55 & 7.51 & 6.51 & 9.21 & 5.71 & 7.47 \\
%     ours & \textbf{6.90} & \textbf{7.40}  & \textbf{7.19} &  &  & \textbf{4.70} &   \\
%     \bottomrule
%   \end{tabular}
%   \caption{Results. Our method demonstrates superior performance.}
%   \label{tab:example}
% \end{table}

\section{Conclusion}
\hspace*{12pt} We propose SNI-SLAM, a semantic SLAM system based on neural implicit representation to improve dense visual mapping and tracking accuracy while providing semantic mapping of the whole scene. We propose feature fusion method based on cross-attention  to enable appearance, geometry, semantic features to potentially promote each other and engage in cross-learning. We propose coarse-to-fine semantic representation to model the semantic information in the scene at multiple levels. This representation can maintain the precision of overall scene semantic information, while considering intricate semantic details. We propose a new decoder design that enables fusion of interpolation features from feature planes, leading to more accurate decoding results.

\noindent\textbf{\small Acknowledgements: }{\small This work was supported in part by the Natural Science Foundation of China under Grant 62225309, 62073222, U21A20480 and 62361166632}

%% file: sec/X_suppl.tex
\clearpage
\setcounter{page}{1}
% \appendix
\maketitle
\section{Overview}
In the supplementary material, the chapters are briefly described as follows:
\begin{itemize}
\item
we present a detailed experimental setup including baseline introduction, implementation details, and evaluation metrics in Sec.~\ref{sec:Experimental Setup}. 
\item
Additional experimental results are given in Sec.~\ref{sec:Additional Experimental Results} to demonstrate the excellent performance of our method. 
\item
We show visualization results in Sec.~\ref{sec:Visualization} on different scenes.
\item 
We give a video demo of real-time mapping and tracking in Sec.~\ref{sec:Video Demo}.
\end{itemize}

\section{Experimental Setup}
\label{sec:Experimental Setup} 
\subsection{Baselines}
To the best of our knowledge, NIDS-SLAM~\cite{haghighi2023neural} is the only existing semantic NeRF-SLAM method. Therefore, we use it as the  baseline for comparing the accuracy of semantic NeRF-based SLAM.
However, NIDS-SLAM~\cite{haghighi2023neural} does not evaluate mesh reconstruction accuracy, so we only compare the metrics of the semantic segmentation accuracy with this method. To provide a more comprehensive baseline for comparison, we also consider Semantic-NeRF~\cite{zhi2021place}, an offline-trained approach that focuses on high-fidelity semantic 3D reconstruction through hours of extensive optimization. 
For SLAM accuracy, we compare our method with state-of-the-art NeRF-based SLAM methods~\cite{imap, nice, yang2022vox, eslam, wang2023co, sandstrom2023point}. Since iMAP~\cite{imap} is not open source, we use iMAP*~\cite{imap} in our experiment, which is the reimplementation of iMAP.

\subsection{Implementation Details}
\noindent\textbf{Hyperparameters.}\hspace*{5pt}
For geometry representation, we adopt the coarse feature planes with a resolution of 24 cm and the fine feature planes with a resolution of 6 cm. For semantic and appearance representation, we employ the coarse feature planes with a resolution of 24 cm and the fine feature planes with a resolution of 3 cm. We use 16-channel feature vectors to represent semantic, geometry, and appearance features for both coarse and fine feature planes, resulting in 32-channel concatenated features input for the decoder. 

For rendering, we first sample \(N_{strat}\) points for each ray by stratified sampling. We then additionally sample \(N_{imp}\) points near surfaces. For pixels with ground truth depths, the \(N_{imp}\) additional points are uniformly sampled within the truncation distance with respect to the depth measurement. For Replica dataset~\cite{straub2019replica}, we set \(N_{strat}=32\) and \(N_{imp}=8\). We conduct 15 optimization iterations for the mapping process and 8 optimization iterations for the tracking process. We sample 4000 pixels for mapping optimization and 2000 pixels for tracking optimization in each iteration. Since ScanNet dataset~\cite{dai2017scannet} scenes are at a larger scale and more complicated, we set \(N_{strat}=48\) and \(N_{imp}=8\). We perform 40 optimization iterations for both the mapping and tracking process. 
 
 We use a window of 5 keyframes for jointly optimizing scene representation, the MLP network, and the camera poses of the selected keyframes. The weighting coefficients of each loss are \(\lambda_{fs}=5\), \(\lambda_{s}=0.1\), \(\lambda_{f}=5\), \(\lambda_{d}=0.1\), \(\lambda_{c}=5\). We use a learning rate of 0.005 for feature planes, 0.001 for the decoder, 0.003 for \(E_{\theta}\), \(H_{\theta}\), and \(F_{\theta}\). For camera pose optimization, we use a learning rate of 0.001 in Replica dataset~\cite{straub2019replica}. In ScanNet ~\cite{dai2017scannet} and TUM RGBD~\cite{sturm2012benchmark} dataset, we use a learning rate of 0.0005 for camera translation optimization and 0.0025 for camera rotation optimization. 

\noindent\textbf{Network details.}\hspace*{5pt}
For the decoder design \(D_{\theta}\), the input geometry features are processed through a two-layer MLP with 16 channels in the hidden layer. 
This MLP outputs 17-channel vectors, where values of the first channel are  used as the output for SDF values. The remaining 16 channels are concatenated with the input semantic features and input appearance features. The concatenated feature vectors are then passed through another two-layer MLP with a hidden size of 16, which outputs the RGB values. We use ReLU activation function for this hidden layer.
Tanh and Sigmoid are respectively used for the output layers of SDF and color values. The input semantic features are processed through a three-layer MLP with 256 channels in the hidden layer to output semantic values.

For cross-attention based feature fusion, the geometry MLP \(E_{\theta}\) is a three-layer MLP while the appearance MLP \(H_{\theta}\) and the fusion MLP \(F_{\theta}\) are both two-layer MLPs. The hidden layers of these MLPs have 16 channels. We use ReLU activation function for the hidden layer of \(H_{\theta}\).

\subsection{Evaluation Metrics}
For mesh reconstruction evaluation, we use \textit{Depth L1 (cm)}, \textit{Accuracy (cm)}, \textit{Completion (cm)}, and \textit{Completion ratio (\%)} with a threshold of 5 cm. We remove unobserved areas outside of any camera radius, as well as additional mesh culling to remove noise points following Co-SLAM~\cite{wang2023co}. 

\textbf{Depth L1 (cm):} the average absolute error between ground truth depth and reconstructed depth. The depth values are generated by randomly sampling 1000 views from both the reconstructed meshes and the ground truth meshes.

\textbf{Accuracy (cm):} the average distance from sampled points on the reconstructed mesh to their nearest ground truth points.

\textbf{Completion (cm):} the average distance from sampled points on the ground truth mesh to their nearest points on the reconstructed mesh.

\textbf{Completion ratio (\%):} the percentage of points in the reconstructed mesh with \textit{Completion} under 5 cm.

For localization accuracy evaluation, we use ATE~\cite{sturm2012benchmark}. including \textit{RMSE (cm)} and \textit{Mean (cm)} metrics. Semantic segmentation is evaluated with respect to mIoU (\%) and per-pixel accuracy (\%)~\cite{long2015fully}. 
\section{Additional Experiments}
\label{sec:Additional Experimental Results}
\begin{figure}
  \centering
  \includegraphics[width=\linewidth]{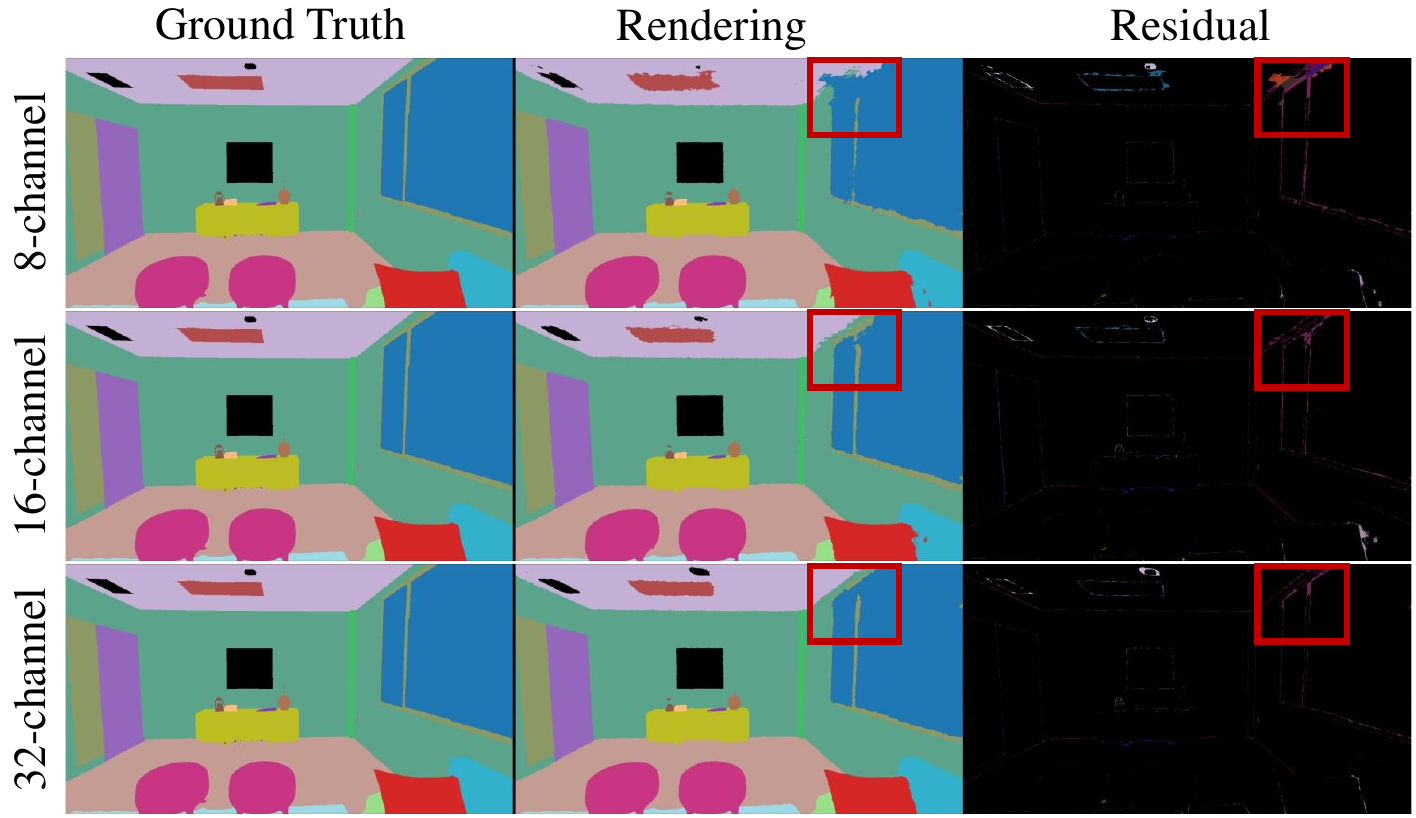}
  \caption{Quantitative analysis on the number of feature channels. With the increase in the number of feature channels, the semantic rendering results achieve higher accuracy.}
  \label{fig:suppl_comp_dim}
\end{figure}

\begin{table}
  \centering
  \resizebox{\columnwidth}{!}{
  \begin{tabular}{l | ccc|c|cc}
    \toprule
    \multirow{2}{*}{Channels} & Reconstruction & Localization & Semantic & \multirow{2}{*}{\#param.~$\downarrow$}& \multicolumn{2}{c}{Runtime}\\
    & Acc. (cm)~$\downarrow$ & RMSE (cm)~$\downarrow$ & mIoU (\%)~$\uparrow$ &  & Map. (ms)~$\downarrow$& Track. (ms)~$\downarrow$ \\
    \midrule  
    Ours-8 & 2.21 & 0.65 & 82.35 & \textbf{3.19} & \textbf{23.6 \texttimes{} 15} & \textbf{7.2 \texttimes{} 8}\\
    Ours-16 & \textbf{2.09} & \textbf{0.50} & 84.50 & 6.23 & 26.9 \texttimes{} 15 & 7.8 \texttimes{} 8\\
    Ours-32 & 2.32 & 1.64 & \textbf{84.65} & 12.37 & 34.6 \texttimes{} 15  & 8.6 \texttimes{} 8 \\
    \toprule
  \end{tabular} }
  \caption{Results of using different feature channel numbers. Mapping and tracking runtime is reported in $ms/iter \times iter$ format. With the increase of channel numbers, the numbers of parameter, mapping and tracking time also increase, as the network requires more time to optimize higher-dimensional features. }
  % At the same time, using higher-dimensional features for scene representation leads to more accurate scene modeling, resulting in higher SLAM accuracy and semantic segmentation accuracy.
  \label{tab:suppl_ablation_dim}
\end{table}

\subsection{More Ablation Studies} 
\noindent\textbf{The number of feature channels.}\hspace*{5pt} 
We conduct an ablation study on the effect of using different numbers of feature channels. 
Fig.~\ref{fig:suppl_comp_dim} demonstrates that as the number of feature channels increases, the semantic rendering results progressively approach the ground truth labels. This can be attributed to the enhanced ability of high-dimensional features to capture and represent more descriptive and diverse information of the environment. 
However, the increase in the number of feature channels leads to higher parameter numbers and longer computation time, as shown in Tab.~\ref{tab:suppl_ablation_dim}. Moreover, using 32-channel features for scene representation with only 15 iterations in mapping and 8 iterations in tracking leads to degraded mapping and tracking performance, as the 32-channel feature planes cannot be sufficiently optimized. Based on the above results, it can be concluded that using 16-channel features in scene representation is a trade-off between the accuracy and runtime.

\begin{figure*}
  \centering
  \includegraphics[width=\linewidth]{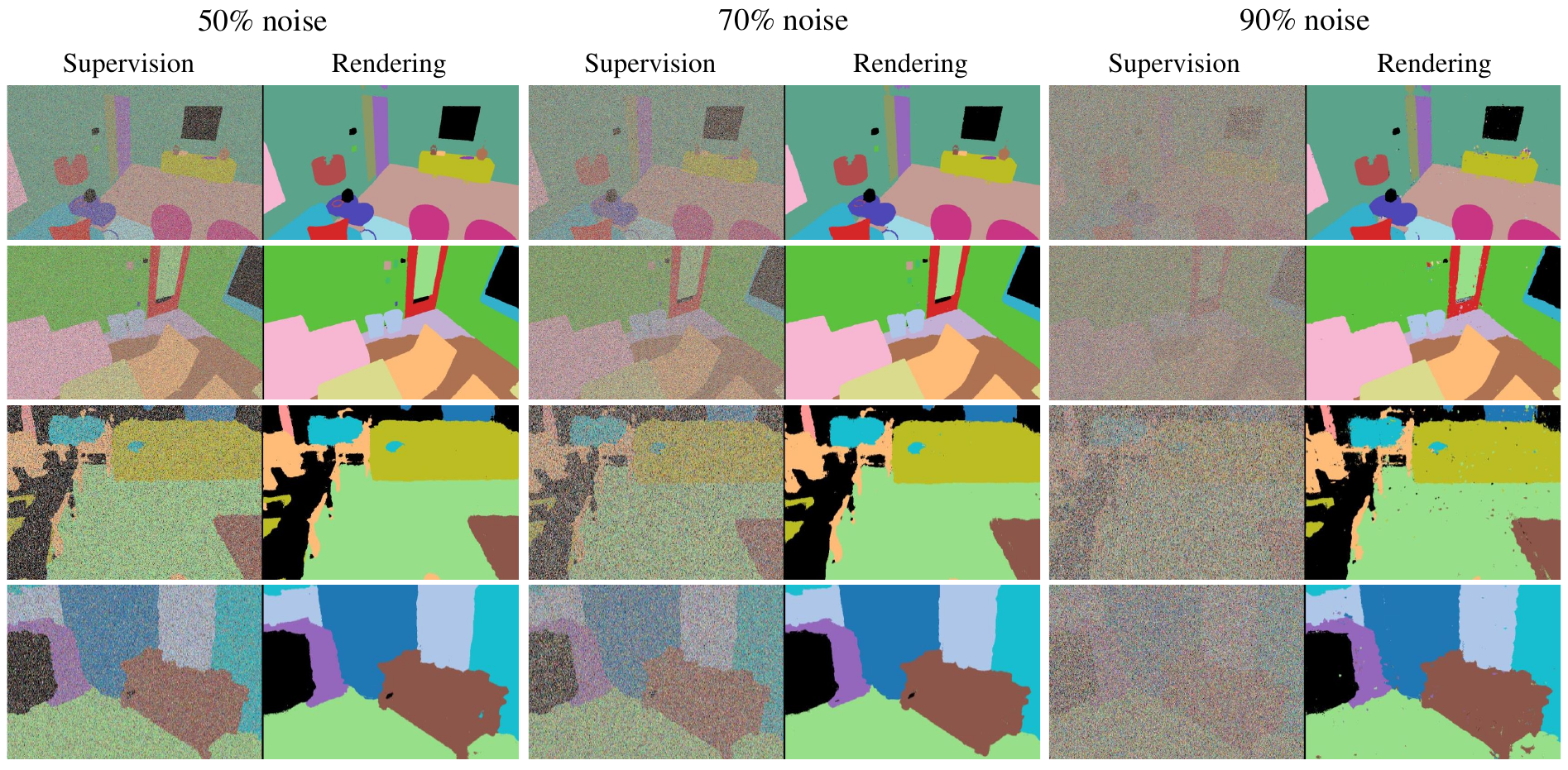}
  \caption{Semantic rendering results with noisy semantic labels for supervision on both Replica~\cite{straub2019replica} and ScanNet~\cite{dai2017scannet} datasets. }
  \label{fig:suppl_noise}
\end{figure*}

\noindent\textbf{Effect of noisy semantic results for supervision.}\hspace*{5pt} Fig.~\ref{fig:suppl_noise} illustrates the robustness of our method when employing noisy semantic supervision signals with various noise ratios. Our approach achieves relatively accurate semantic rendering results even when using noisy semantic supervision with noise ratios up to 90\% in both Replica~\cite{straub2019replica} and ScanNet~\cite{dai2017scannet} datasets. 
This robustness is attributable to our feature collaboration strategy, which effectively fuses geometric and appearance information. Specifically, geometric features offer information about the shapes and spatial relationships of objects. Such information can mitigate the effects of inconsistencies caused by noisy semantic labels, primarily because geometric attributes typically maintain their stability despite the presence of semantically noisy labels. Moreover, appearance features provide detailed surface characteristics of objects, such as texture and color, which are crucial for differentiating similar objects with distinct identities even when their semantic labels may be incorrect. 
Through the effective collaboration of different features, a more reliable and comprehensive understanding of the scene is achieved. Therefore, we are capable of reducing the negative effects of high-noise labels in the semantic supervision. 

\begin{table}
  \centering
  \resizebox{\columnwidth}{!}{
  \begin{tabular}{l | ccc | cc}
    \toprule
    \multirow{2}{*}{Name} & \multicolumn{3}{c|}{Reconstruction [cm]} & \multicolumn{2}{c}{Localization [cm]} \\
    & Acc.~$\downarrow$ & Comp.~$\downarrow$ & Comp.Ratio(\%) $\uparrow$ & RMSE~$\downarrow$ & Mean~$\downarrow$\\
    \midrule  
    w/o semantics & 1.83 & 1.78 & 96.07 & 0.58 & 0.52\\
    w/ semantics & \textbf{1.66} & \textbf{1.56} & \textbf{96.74} & \textbf{0.55} & \textbf{0.48}\\
    \bottomrule
  \end{tabular} }
  \caption{Ablation study of adding semantic information. Semantic information can enhance the expression of geometry and appearance, achieving higher accuracy in mapping and tracking.}
  \label{tab:suppl_ablation_sem}
\end{table}

\noindent\textbf{Effect of whether to use semantic information.}\hspace*{5pt}
Tab.~\ref{tab:suppl_ablation_sem} demonstrates that semantic information can enhance the expression of geometric and appearance information, leading to higher accuracy in mesh reconstruction and localization. 
Specifically, semantic information provides a richer and more detailed context for the interpretation of geometric and appearance information.
For example, knowing the semantic segmentation results of a set of geometric data corresponding to a chair can guide the model to reconstruct and locate the object more accurately and realistically. 
Therefore, it can be concluded that the integration of semantic information offers a more comprehensive understanding of the environment.

\noindent\textbf{Effect of whether to add Bundle Adjustment (BA).}\hspace*{5pt}
As shown in Tab.~\ref{tab:ba}, we incorporate BA into our SNI-SLAM following BARF~\cite{lin2021barf} and L2G-NeRF~\cite{chen2023local}. Such integration only shows a slight improvement in SLAM accuracy because current BA NeRFs lack adequate constraints for semantic SLAM optimization.

\noindent\textbf{Ablation of our innovations across different scenes.}\hspace*{5pt}
We conduct ablation experiments across 8 scenes in Tab.~\ref{tab:ablation_room01}-\ref{tab:ablation_office34}, noticing that each innovation is added from baseline. All innovations achieve large increased accuracy (average 13\% improvement). 
Specifically, cross-attention based feature fusion and feature loss increase semantic accuracy by 23.2\%, attributed to mutual reinforcement between different modality features of scene and higher-level supervision for semantic optimization. Moreover, one-way correlation decoder can achieve up to 33\% improvement in metrics.

\begin{table}
  \centering
  \resizebox{\linewidth}{!}{%
  \begin{tabular}{l | ccc | c | c}
    \toprule
    Name & Acc.[cm]\downarrow & Comp.[cm]\downarrow & Comp.Ratio(\%)\uparrow & RMSE[cm]\downarrow & mIoU(\%)\uparrow\\
    \midrule
    w/o BA & 2.520 & \textbf{1.873} & 95.201 & 0.320 & 85.91\\
    w/ BA & \textbf{2.517} & 1.876 & \textbf{95.350} & \textbf{0.312} & \textbf{85.95}\\
    \bottomrule
  \end{tabular}%
  }
  \caption{Experiment of whether to use BA on Replica~\cite{straub2019replica}.}
  \label{tab:ba}
\end{table}

\subsection{Semantic Segmentation Results}
Tab.~\ref{tab:replica_semantic} shows per-scene semantic segmentation results of both online and offline 3D semantic reconstruction methods. For online methods, NIDS-SLAM~\cite{haghighi2023neural} and our SNI-SLAM, take RGB-D frames as input and perform real-time mapping and camera pose estimation, which require \textbf{several minutes} for semantic scene reconstruction. For the offline method, Semantic-NeRF~\cite{zhi2021place}, takes camera pose and RGB-D frames as input and requires nearly \textbf{10 hours} of training to obtain semantic reconstruction results, which is several tens of times the duration required by the online methods.
Our method outperforms the online method NIDS-SLAM~\cite{haghighi2023neural} of all scenes and all metrics. 
Compared with the offline method Semantic-NeRF~\cite{zhi2021place}, our real-time semantic mapping method achieves similar results in pixel accuracy metric.

\subsection{Reconstruction and Localization Results} 
Tab.~\ref{tab:replica_slam_all} demonstrates per-scene quantitative evaluation of our method with existing NeRF-based SLAM method in Replica dataset~\cite{straub2019replica}. Our method achieves the best performance in \textit{Depth L1}, \textit{Completion}, \textit{Completion ratio (\%)}, ATE RMSE, and ATE Mean metrics across all scenes. In some scenes, our method can improve localization accuracy by up to 52\% and reconstruction accuracy by up to 32\%, demonstrating excellent performance across scenes. This remarkable improvement is attributed  to the thoughtful design of the semantic NeRF-based SLAM framework.

\begin{table*}
  \centering
  \vspace{-11pt}
  \resizebox{2.0\columnwidth}{!}{
  \begin{tabular}{l|l| ccc | ccc}
    \toprule
    & \multirow{2}{*}{Name} & \multicolumn{3}{c|}{room0} & \multicolumn{3}{c}{room1} \\
    & & Acc.[cm]$\downarrow$ & RMSE[cm]$\downarrow$ & mIoU(\%)$\uparrow$ & Acc.[cm]$\downarrow$ & RMSE[cm]$\downarrow$ & mIoU(\%)$\uparrow$\\
    \midrule
    (a) & Ours (w/o all innovations in ablation studies, \bf{baseline model}) 
    & 2.45 & 0.70 & 65.12 
    & 2.08 & 1.13 & 54.60\\
    \midrule
    (b) & Ours ($+$ feature loss to (a))  
    & 2.20 & 0.55 & 71.30 
    & 1.96 & 0.92 & 60.53 \\
    (c) & Ours ($+$ feature fusion with feature loss to (a)) 
    & 2.13 & 0.53 & 75.31   
    & 1.85 & 0.67 & 67.29 \\
    (d) & Ours ($+$ decoder design to (a)) 
    & 2.21 & 0.53 & 72.56
    & 1.82 & 0.89 & 58.91\\
    (e) & Ours ($+$ hierarchical semantic representation to (a)) 
    & 2.30 & 0.55 & 80.24
    & 1.92 & 0.79 & 75.32\\
     \midrule
    (f) & Ours (full, with all innovations in ablation studies, \bf{best model})
    & \bf{2.09} & \bf{0.50} & \bf{87.32}
    & \bf{1.66} & \bf{0.55} & \bf{86.33}\\
    \bottomrule
  \end{tabular}}
   \caption{Ablation study of our innovations on \textbf{room0} and \textbf{room1} of Replica~\cite{straub2019replica}.}
  \label{tab:ablation_room01}
\end{table*}

\begin{table*}
  \centering
  \resizebox{2.0\columnwidth}{!}{
  \begin{tabular}{l|l| ccc | ccc}
    \toprule
    & \multirow{2}{*}{Name} & \multicolumn{3}{c|}{room2} & \multicolumn{3}{c}{office0} \\
    & & Acc.[cm]$\downarrow$ & RMSE[cm]$\downarrow$ & mIoU(\%)$\uparrow$ & Acc.[cm]$\downarrow$ & RMSE[cm]$\downarrow$ & mIoU(\%)$\uparrow$\\
    \midrule
    (a) & Ours (w/o all innovations in ablation studies, \bf{baseline model}) 
    & 1.77 & 0.55 & 63.91  
    & 1.59 & 0.83 & 71.52\\
    \midrule
    (b) & Ours ($+$ feature loss to (a))  
    & 1.73 & 0.50 & 72.44 
    & 1.50 & 0.59 & 74.23 \\
    (c) & Ours ($+$ feature fusion with feature loss to (a)) 
    & 1.70 & 0.46 & 75.38
    & 1.48 & 0.55 & 77.74 \\
    (d) & Ours ($+$ decoder design to (a)) 
    & 1.65 & 0.47 & 74.28   
    & 1.49 & 0.56 & 75.94 \\
    (e) & Ours ($+$ hierarchical semantic representation to (a)) 
    & 1.71 & 0.51 & 82.01   
    & 1.51 & 0.55 & 84.12\\
     \midrule
    (f) & Ours (full, with all innovations in ablation studies, \bf{best model})
    & \bf{1.64} & \bf{0.45} & \bf{85.16}
    & \bf{1.46} & \bf{0.33} & \bf{86.01}\\
    \bottomrule
  \end{tabular} }
   \caption{Ablation study of our innovations on \textbf{room2} and \textbf{office0} of Replica~\cite{straub2019replica}.}
   \vspace{-0.1in}
  \label{tab:ablation_room2}
\end{table*}

\begin{table*}
  \centering
  \resizebox{2.0\columnwidth}{!}{
  \begin{tabular}{l|l| ccc | ccc}
    \toprule
    & \multirow{2}{*}{Name} & \multicolumn{3}{c|}{office1} & \multicolumn{3}{c}{office2} \\
    & & Acc.[cm]$\downarrow$ & RMSE[cm]$\downarrow$ & mIoU(\%)$\uparrow$ & Acc.[cm]$\downarrow$ & RMSE[cm]$\downarrow$ & mIoU(\%)$\uparrow$\\
    \midrule
    (a) & Ours (w/o all innovations in ablation studies, \bf{baseline model}) 
    & 1.84 & 0.89 & 58.65
    & 2.99 & 0.49 & 64.36\\
    \midrule
    (b) & Ours ($+$ feature loss to (a))  
    & 1.65 & 0.74 & 64.79   
    & 2.74 & 0.43 & 67.62 \\
    (c) & Ours ($+$ feature fusion with feature loss to (a)) 
    & 1.62 & 0.58 & 67.35      
    & 2.70 & 0.41 & 72.01\\
    (d) & Ours ($+$ decoder design to (a)) 
    & 1.64 & 0.73 & 62.92   
    & 2.88 & 0.46 & 68.45\\
    (e) & Ours ($+$ hierarchical semantic representation to (a)) 
    & 1.80 & 0.65 & 68.14   
    & 2.79 & 0.45 & 75.18\\
     \midrule
    (f) & Ours (full, with all innovations in ablation studies, \bf{best model})
    & \bf{1.58} & \bf{0.41} & \bf{78.13}
    & \bf{2.52} & \bf{0.32} & \bf{85.91}\\
    \bottomrule
  \end{tabular}}
   \caption{Ablation study of our innovations on \textbf{office1} and \textbf{office2} of Replica~\cite{straub2019replica}.}
  \label{tab:ablation_office12}
\end{table*}

\begin{table*}
  \centering
  \resizebox{2.0\columnwidth}{!}{
  \begin{tabular}{l|l| ccc | ccc}
    \toprule
    & \multirow{2}{*}{Name} & \multicolumn{3}{c|}{office3} & \multicolumn{3}{c}{office4} \\
    & & Acc.[cm]$\downarrow$ & RMSE[cm]$\downarrow$ & mIoU(\%)$\uparrow$ & Acc.[cm]$\downarrow$ & RMSE[cm]$\downarrow$ & mIoU(\%)$\uparrow$\\
    \midrule
    (a) & Ours (w/o all innovations in ablation studies, \bf{baseline model}) 
    & 2.78 & 0.72 & 52.42  
    & 2.19 & 0.65 & 61.92\\
    \midrule
    (b) & Ours ($+$ feature loss to (a))  
    & 2.57 & 0.66 & 58.71   
    & 2.13 & 0.59 & 66.68\\
    (c) & Ours ($+$ feature fusion with feature loss to (a)) 
    & 2.55 & 0.63 & 61.39    
    & 2.09 & 0.57 & 68.51\\
    (d) & Ours ($+$ decoder design to (a)) 
    & 2.53 & 0.63 & 58.88  
    & 2.14 & 0.63 & 69.25 \\
    (e) & Ours ($+$ hierarchical semantic representation to (a)) 
    & 2.59 & 0.67 & 64.02   
    & 2.17 & 0.62 & 74.96 \\
     \midrule
    (f) & Ours (full, with all innovations in ablation studies, \bf{best model})
    & \bf{2.51} & \bf{0.62} & \bf{73.41}
    & \bf{2.07} & \bf{0.47} & \bf{79.32}\\
    \bottomrule
  \end{tabular}}
   \caption{Ablation study of our innovations on \textbf{office3} and \textbf{office4} of Replica~\cite{straub2019replica}.}
  \label{tab:ablation_office34}
\end{table*}

\begin{table*}
  \centering
  \resizebox{\textwidth}{!}{
  \begin{tabular}{lc|cc|cc|cc|cc|cc|cc|cc|cccc}
    \toprule
    & \multirow{2}{*}{Methods} & \multicolumn{2}{c|}{room0}  & \multicolumn{2}{c|}{room1}  & \multicolumn{2}{c|}{room2}  & \multicolumn{2}{c|}{office0}  & \multicolumn{2}{c|}{office1} & \multicolumn{2}{c|}{office2} & \multicolumn{2}{c|}{office3} & \multicolumn{2}{c}{office4}\\
    & & Acc.  & mIoU  & Acc.  & mIoU  & Acc.  & mIoU  & Acc.  & mIoU  & Acc.  & mIoU  & Acc.  & mIoU  & Acc.  & mIoU  & Acc.  & mIoU  \\
    \midrule
    Offline & Semantic-NeRF~\cite{zhi2021place} & 98.34 & 97.00 & 98.71 & 97.28 & 97.59 & 95.92 & 98.67 & 97.44 & 97.07 & 93.48 & 97.62 & 92.42 & 96.74 & 94.31 & 96.62 & 94.22\\
    \midrule
    \multirow{2}{*}{Online} & NIDS-SLAM~\cite{haghighi2023neural} & 97.76 &82.45 & 98.50 & 84.08& 98.76 & 76.99& 98.89 & 85.94& -- & -- & -- & -- & -- & -- & -- & --\\
    & SNI-SLAM (Ours) & \textbf{98.53} & \textbf{88.42}& \textbf{98.61} & \textbf{87.43} & \textbf{98.80} & \textbf{86.16} & \textbf{99.17} & \textbf{87.63} & \textbf{99.18} & \textbf{78.63} & \textbf{99.13} & \textbf{86.49} & \textbf{98.66} & \textbf{74.01}& \textbf{99.06} & \textbf{80.22}\\
    
    \bottomrule
  \end{tabular}}
  \caption{Quantitative comparison of SNI-SLAM with existing semantic NeRF-based SLAM method NIDS-SLAM~\cite{haghighi2023neural} and offline 3D semantic reconstruction method Semantic-NeRF~\cite{zhi2021place}. For a fair comparison, the results are obtained using ground truth semantic labels for supervision. 
  Online methods only take RGB-D frames as input, while offline method requires RGB-D frames and corresponding camera poses. Online methods perform scene reconstruction in several minutes while offline method require hours of training.  Our method surpasses the performance of NIDS-SLAM~\cite{haghighi2023neural} and achieves comparable results with Semantic-NeRF~\cite{zhi2021place} in pixel accuracy metric.}
  \label{tab:replica_semantic}
\end{table*}
\begin{table*}
    \centering
    \small
    \begin{tabular}{c|c|cccc|cc}
    \toprule
    & \multirow{2}{*}{Methods} & \multicolumn{4}{c|}{Reconstruction [cm]} & \multicolumn{2}{c}{Localization [cm]}    \\
    && Depth L1 $\downarrow$ & Acc. $\downarrow$ & Comp. $\downarrow$ & Comp.Ratio(\%) $\uparrow$ & RMSE $\downarrow$ & Mean $\downarrow$ \\                      
    \midrule
    \multirow{6}{*}{office0}                  
    &iMAP*~\cite{imap} & 3.79 & 3.34 & 3.62 & 83.59 & 3.31 & 2.74\\
    &NICE-SLAM~\cite{nice} & 1.43 & 1.83 & 1.84 & 94.93 & 1.50 & 1.32\\
    &Vox-Fusion~\cite{yang2022vox} & 3.44 & 1.63 & 1.87 & 93.86 & 1.35 & 0.98\\
    &Co-SLAM~\cite{wang2023co} &  \cellcolor{third}1.24 & \cellcolor{second}1.57 & \cellcolor{third}1.56 & \cellcolor{third}96.09 & \cellcolor{third}0.69 & \cellcolor{third}0.63\\
    &ESLAM~\cite{eslam} & \cellcolor{second}0.71 & \cellcolor{third}1.61 & \cellcolor{second}1.45 & \cellcolor{second}98.45 & \cellcolor{second}0.61 & \cellcolor{second}0.45\\
    &SNI-SLAM (Ours) &  \cellcolor{first} \textbf{0.55} & \cellcolor{first} \textbf{1.46} & \cellcolor{first}\textbf{1.30} & \cellcolor{first}\textbf{98.70} & \cellcolor{first}\textbf{0.33} & \cellcolor{first}\textbf{0.28}\\

    \midrule
    \multirow{6}{*}{office1}                       
    &iMAP*~\cite{imap}  & 3.76 & 2.10 & 3.62 & 88.45 & 1.42 & 1.15 \\
    &NICE-SLAM~\cite{nice} & 1.58 & 1.76 & 1.82 & 94.11 & 1.01 & 0.91 \\
    &Vox-Fusion~\cite{yang2022vox} & 1.77 & \cellcolor{third}1.60 & 1.66 & 94.40 & 1.76 & 1.29\\
    &Co-SLAM~\cite{wang2023co} & \cellcolor{third}1.48 & \cellcolor{first}\textbf{1.31} & \cellcolor{third}1.59 & \cellcolor{third}94.65 & \cellcolor{second}0.56 & \cellcolor{third}0.52\\
    &ESLAM~\cite{eslam} & \cellcolor{second}1.02 & 1.82 & \cellcolor{second}1.30 & \cellcolor{second}97.60 & \cellcolor{third}0.59 & \cellcolor{second}0.51\\
    &SNI-SLAM (Ours) & \cellcolor{first}\textbf{0.97} & \cellcolor{second}1.58 & \cellcolor{first}\textbf{1.26} & \cellcolor{first}\textbf{97.70} & \cellcolor{first}\textbf{0.41} & \cellcolor{first}\textbf{0.35}\\
	
    \midrule
    \multirow{6}{*}{office2}                  
    &iMAP*~\cite{imap}  & 3.97 & 4.06 & 4.73 & 79.73 & 7.17 & 4.81 \\
    &NICE-SLAM~\cite{nice} & 2.70 & 3.18 & 3.11 & 88.27 & 1.85 & 1.51\\
    &Vox-Fusion~\cite{yang2022vox} & 3.52 & \cellcolor{first} \textbf{2.02} & 3.03 & 88.94 & \cellcolor{third}1.18 & \cellcolor{third}0.73\\
    &Co-SLAM~\cite{wang2023co} & \cellcolor{third}1.86 & \cellcolor{third}2.84 & \cellcolor{third}2.43 & \cellcolor{third}91.63 & 2.12 & 1.98\\
    &ESLAM~\cite{eslam} & \cellcolor{second}0.93 & 2.95 & \cellcolor{second}1.92 & \cellcolor{second}95.07 & \cellcolor{second}0.67 & \cellcolor{second}0.50\\
    &SNI-SLAM (Ours) & \cellcolor{first} \textbf{0.89} & \cellcolor{second}2.52 & \cellcolor{first} \textbf{1.87} & \cellcolor{first} \textbf{95.20} & \cellcolor{first} \textbf{0.32} & \cellcolor{first} \textbf{0.28}\\	
		
    \midrule
    \multirow{6}{*}{office3}                    
    &iMAP*~\cite{imap}  & 5.61 & 4.20 & 5.49 & 73.90 & 6.32 & 4.89 \\
    &NICE-SLAM~\cite{nice} & 2.10 & 3.01 & 3.16 & 87.68 & 5.67 & 2.53\\
    &Vox-Fusion~\cite{yang2022vox} & 1.82 & \cellcolor{first} \textbf{2.33} & 2.81 & 89.10 & \cellcolor{third}1.11 & \cellcolor{third}0.69\\
    &Co-SLAM~\cite{wang2023co} & \cellcolor{third}1.66 & 3.06 & \cellcolor{third}2.72 & \cellcolor{third}90.72 & 1.62 & 1.47\\
    &ESLAM~\cite{eslam} & \cellcolor{second}1.03 & \cellcolor{third}2.55 & \cellcolor{second}2.20 & \cellcolor{second}95.05 & \cellcolor{second}0.74 & \cellcolor{second}0.64\\
    &SNI-SLAM (Ours) & \cellcolor{first} \textbf{0.75} & \cellcolor{second}2.51 & \cellcolor{first} \textbf{2.07} & \cellcolor{first} \textbf{95.40} & \cellcolor{first} \textbf{0.62} & \cellcolor{first} \textbf{0.56}\\
						
    \midrule
    \multirow{6}{*}{office4}
    &iMAP*~\cite{imap}  & 5.71 & 4.34 & 6.65 & 74.77 & 2.55 & 2.10 \\
    &NICE-SLAM~\cite{nice} & 2.06 & 2.54 & 3.61 & 87.23 & 3.53 & 2.52\\
    &Vox-Fusion~\cite{yang2022vox} & 4.84 & \cellcolor{first} \textbf{2.02} & 3.51 & 86.53 & 1.64 & 1.18\\
    &Co-SLAM~\cite{wang2023co} & \cellcolor{third}1.54 & 2.23 & \cellcolor{third}2.52 & \cellcolor{third}90.44 & \cellcolor{third}0.87 & \cellcolor{third}0.68\\
    &ESLAM~\cite{eslam} & \cellcolor{second}1.18 & \cellcolor{third}2.10 & \cellcolor{second}2.13 & \cellcolor{second}94.31 & \cellcolor{second}0.66 & \cellcolor{second}0.54\\
    &SNI-SLAM (Ours) & \cellcolor{first} \textbf{0.97} & \cellcolor{second}2.07 & \cellcolor{first} \textbf{2.10} & \cellcolor{first} \textbf{94.40} & \cellcolor{first} \textbf{0.47} & \cellcolor{first} \textbf{0.40}\\

    \midrule
    \multirow{6}{*}{room0}                     
    &iMAP*~\cite{imap}  & 5.08 & 4.01 & 5.84 & 78.34 & 6.33 & 3.85 \\
    &NICE-SLAM~\cite{nice} & 1.79 & 2.44 & 2.60 & 91.81 & 1.86 & 1.49\\
    &Vox-Fusion~\cite{yang2022vox} & 1.76 & \cellcolor{first}\textbf{1.77} & 2.69 & 92.03 & 1.37 & 1.03\\
    &Co-SLAM~\cite{wang2023co} & \cellcolor{third}1.05 & \cellcolor{third}2.11 & \cellcolor{third}2.02 & \cellcolor{third}95.26 & \cellcolor{second}0.72 & \cellcolor{second}0.57\\
    &ESLAM~\cite{eslam} & \cellcolor{second}0.73 & 2.15 & \cellcolor{second}1.79 & \cellcolor{second}97.39 & \cellcolor{third}0.84 & \cellcolor{third}0.67\\
    &SNI-SLAM (Ours) & \cellcolor{first} \textbf{0.55} & \cellcolor{second} 2.09 & \cellcolor{first} \textbf{1.73} & \cellcolor{first} \textbf{97.80} & \cellcolor{first} \textbf{0.50} & \cellcolor{first} \textbf{0.43}\\    
	
    \midrule
    \multirow{6}{*}{room1}  
    &iMAP*~\cite{imap}  & 3.44 & 3.04 & 4.40 & 85.85 & 3.46 & 2.91 \\
    &NICE-SLAM~\cite{nice} & 1.33 & 2.10 & 2.19 & 93.56 & 2.37 & 1.92\\
    &Vox-Fusion~\cite{yang2022vox} & 2.52 & \cellcolor{first} \textbf{1.51} & 2.31 & 92.47 & 1.90 & 1.35\\
    &Co-SLAM~\cite{wang2023co} & \cellcolor{third}0.85 & \cellcolor{third}1.68 & \cellcolor{third}1.81	& \cellcolor{third}95.19 & \cellcolor{third}0.85 & \cellcolor{third}0.73\\
    &ESLAM~\cite{eslam} & \cellcolor{second}0.74 & 1.94 & \cellcolor{second}1.58 & \cellcolor{second}96.50 & \cellcolor{second}0.72 & \cellcolor{second}0.58\\
    &SNI-SLAM (Ours) & \cellcolor{first} \textbf{0.58} & \cellcolor{second}1.66 & \cellcolor{first} \textbf{1.56} & \cellcolor{first} \textbf{96.74} & \cellcolor{first} \textbf{0.55 }& \cellcolor{first} \textbf{0.48}\\
							
    \midrule
    \multirow{6}{*}{room2}     
    &iMAP*~\cite{imap}  & 5.78 & 3.84 & 5.07 & 79.40 & 2.65 & 2.50 \\
    &NICE-SLAM~\cite{nice} & 2.20 & 2.17 & 2.73 & 91.48 & 2.26 & 1.65\\
    &Vox-Fusion~\cite{yang2022vox} & 3.58 & 2.23 & 2.58 & 90.13 & 1.47 & 1.02\\
    &Co-SLAM~\cite{wang2023co} & \cellcolor{third}2.37 & \cellcolor{third}1.99 & \cellcolor{third}1.96 & \cellcolor{third}93.58 & \cellcolor{third}1.02 & \cellcolor{third}0.87\\
    &ESLAM~\cite{eslam} & \cellcolor{second}1.26 & \cellcolor{second}1.68 & \cellcolor{second}1.65 & \cellcolor{second}96.99 & \cellcolor{second}0.53 & \cellcolor{second}0.44\\
    &SNI-SLAM (Ours) & \cellcolor{first} \textbf{0.87} & \cellcolor{first} \textbf{1.64} & \cellcolor{first} \textbf{1.62} & \cellcolor{first} \textbf{97.05} & \cellcolor{first} \textbf{0.45} & \cellcolor{first} \textbf{0.38}\\    
		
    \bottomrule
    \end{tabular}
    \caption{Per-scene mesh reconstruction and localization accuracy results in Replica dataset~\cite{straub2019replica}. Best results are highlighted as \colorbox{first}{\textbf{first}}, second best results are highlighted as \colorbox{second}{second}. Our method achieves state-of-the-art performance in all scenes.}
    \label{tab:replica_slam_all}
\end{table*}

\section{Visualization} 
\label{sec:Visualization}
\begin{figure*}
  \centering
  \includegraphics[width=\linewidth]{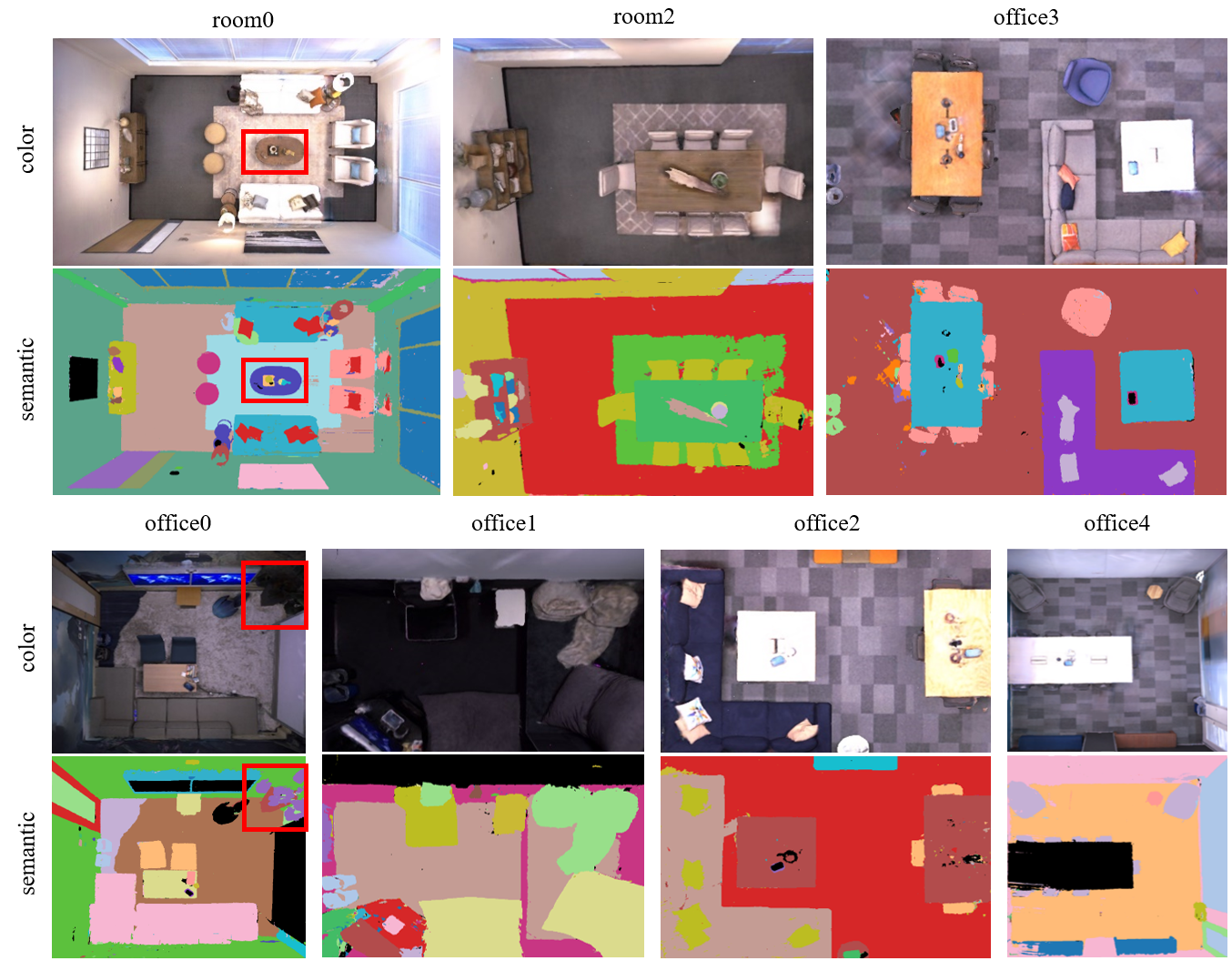}
  \caption{Top-view semantic reconstruction results of SNI-SLAM on the Replica dataset~\cite{straub2019replica}. Our method is capable of achieving relatively accurate results through optimization during the real-time mapping process. As shown in \textcolor{red}{red colored box} of room0, small objects on the table are segmented accurately. \textcolor{red}{Red colored box} of office0 displays that flowers with complex shapes can be precisely segmented.}
  \label{fig:top_view}
\end{figure*}

We demonstrate top-view semantic mapping results in Fig.~\ref{fig:top_view}, showing that our method is capable of achieving accurate segmentation results, even for objects with complex shapes, such as flowers, and small objects on the table. This capability is due to the hierarchical semantic representation, which provides a coarse-to-fine level of understanding. This approach initially identifies the overall semantic information of the scene, then gradually refines this understanding to capture intricate details, even for small or complex objects. Such representation allows for a more comprehensive perception of the environment, facilitating accurate segmentation results.

\begin{figure*}
  \centering
  \includegraphics[width=\linewidth]{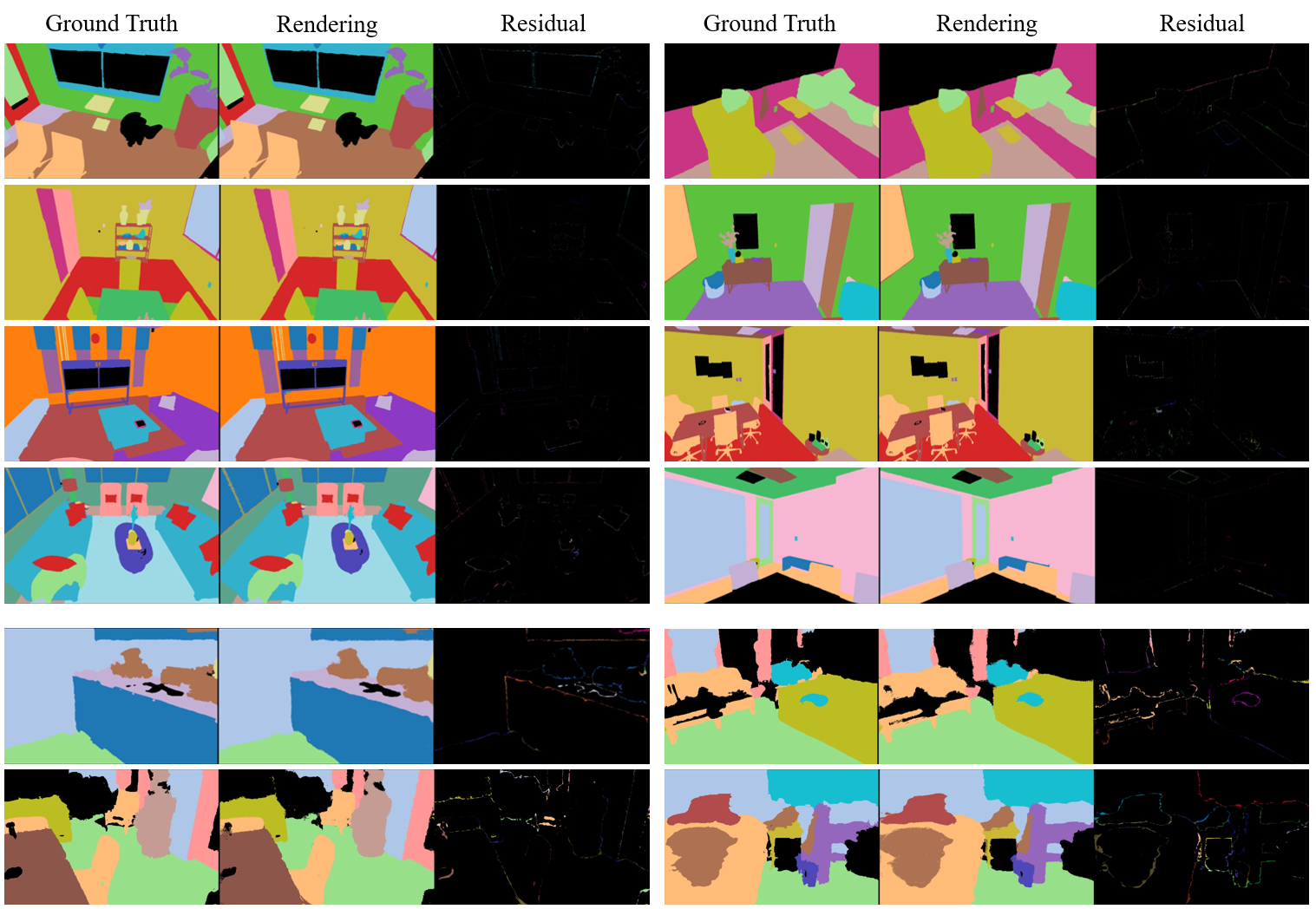}
  \caption{Semantic rendering results of SNI-SLAM on Replica~\cite{straub2019replica} and ScanNet~\cite{dai2017scannet} datasets. Residual visualizes the difference between rendering results and the ground truth labels. It can be observed that our method can achieve excellent semantic segmentation accuracy.}
  \label{fig:suppl_sem_render}
\end{figure*}

\begin{figure*}
  \centering
  \includegraphics[width=\linewidth]{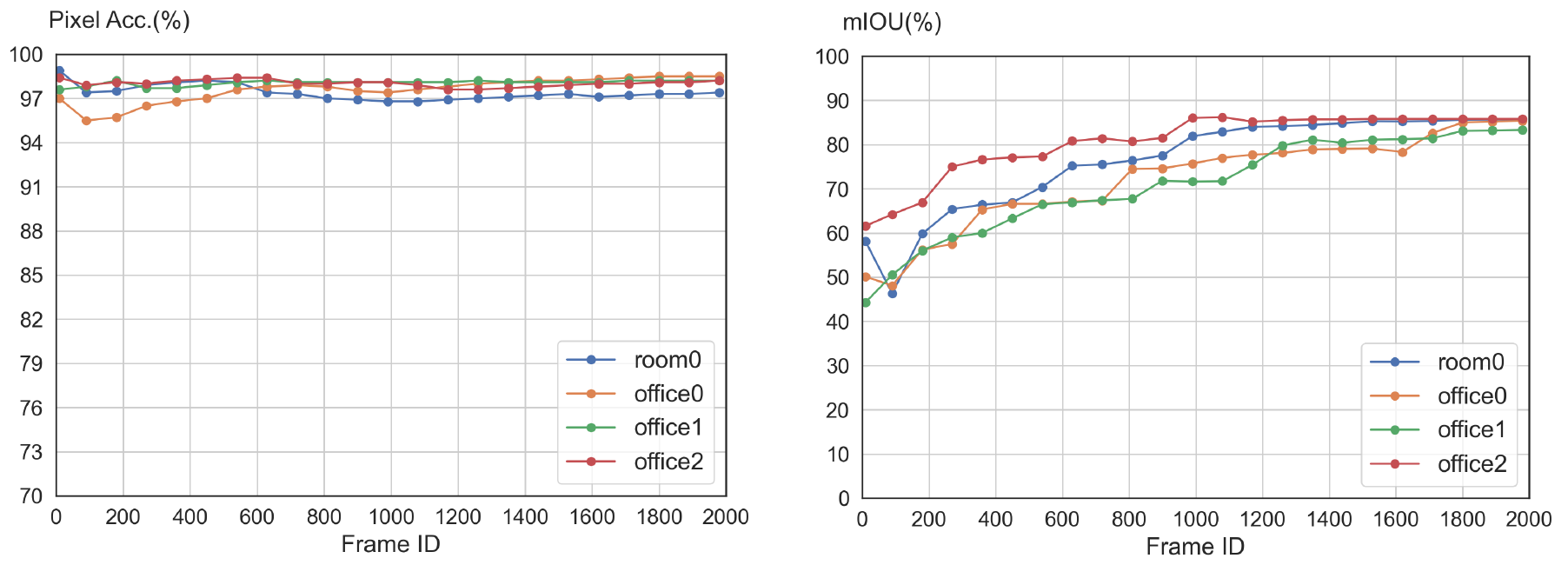}
    \vspace{-0.2in}
  \caption{Qualitative results of SNI-SLAM in real-time semantic mapping from the first frame to the last frame. The y-axis represents pixel accuracy and mIOU of rendering labels, the x-axis represents frame index of simultaneously mapping and tracking. The graph displays that our method can already achieve high accuracy at the beginning of the mapping.}
  \label{fig:acc_miou}
\end{figure*}

Semantic rendering results are shown in Fig.~\ref{fig:suppl_sem_render}. From the residual, we can observe that our method achieves excellent semantic segmentation accuracy in both Replica~\cite{straub2019replica} and ScanNet~\cite{dai2017scannet} datasets. Fig.~\ref{fig:acc_miou} shows pixel accuracy and mIoU changing curve of semantic rendering result from the first frame to the last frame in real-time semantic mapping and tracking. This graph illustrates the fast convergence speed of SNI-SLAM  as well as the high semantic accuracy of real-time mapping, which is attributed to loss construction at the feature level. Such high-level guidance for scene optimization is able to accelerate convergence speed as well as achieving accurate semantic mapping.

\begin{figure*}
  \centering
  \includegraphics[width=\linewidth]{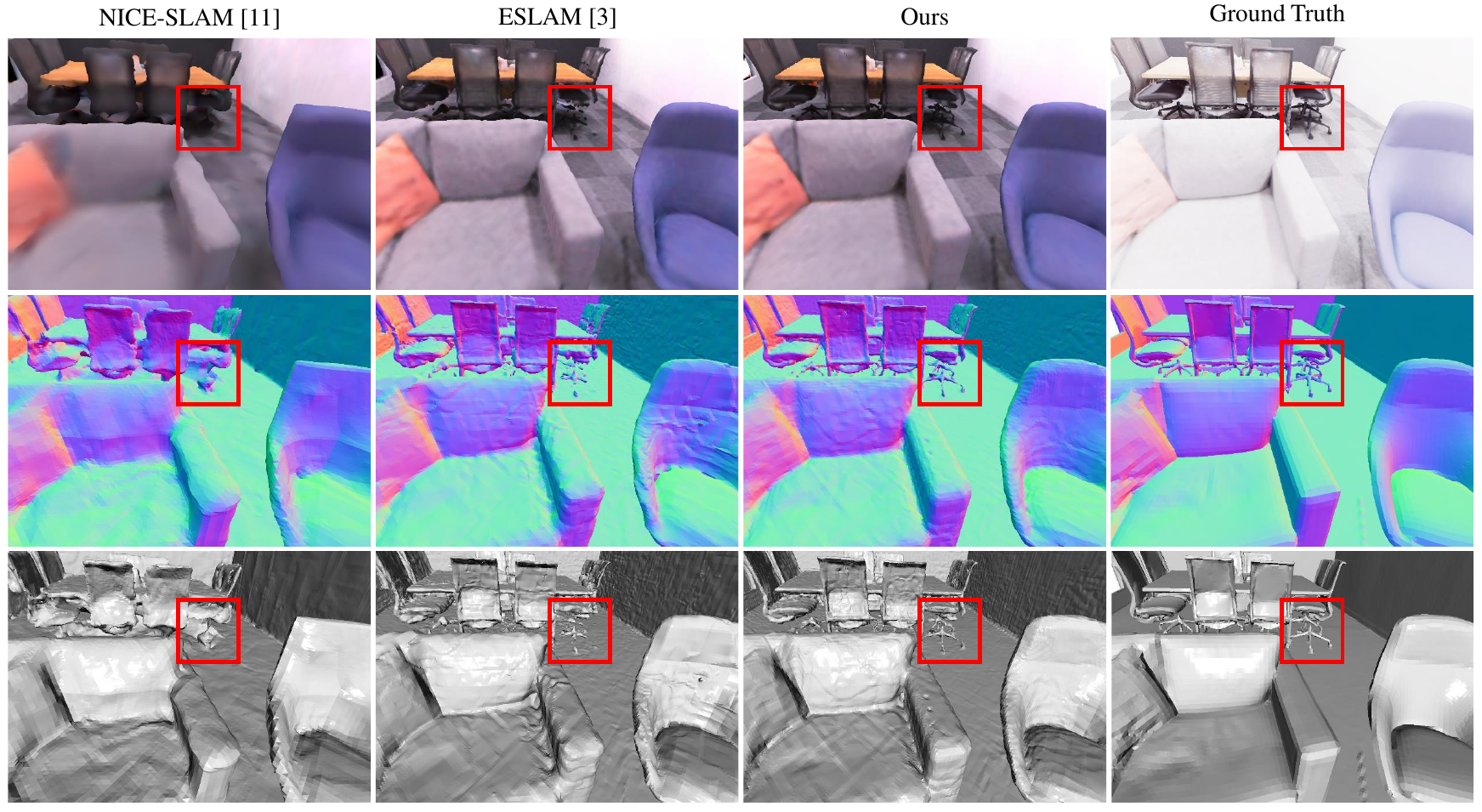}
  \caption{Qualitative comparison of our method with existing NeRF-based SLAM methods on Replica~\cite{straub2019replica} of office3 using different shading mode. As shown in \textcolor{red}{red colored box}, other methods cannot accurately model chair legs while our method can. Moreover, our method achieves more accurate surface reconstruction results than baseline.}
  \label{fig:suppl_off3_detail}
\end{figure*}

\begin{figure*}
  \centering
  \includegraphics[width=\linewidth]{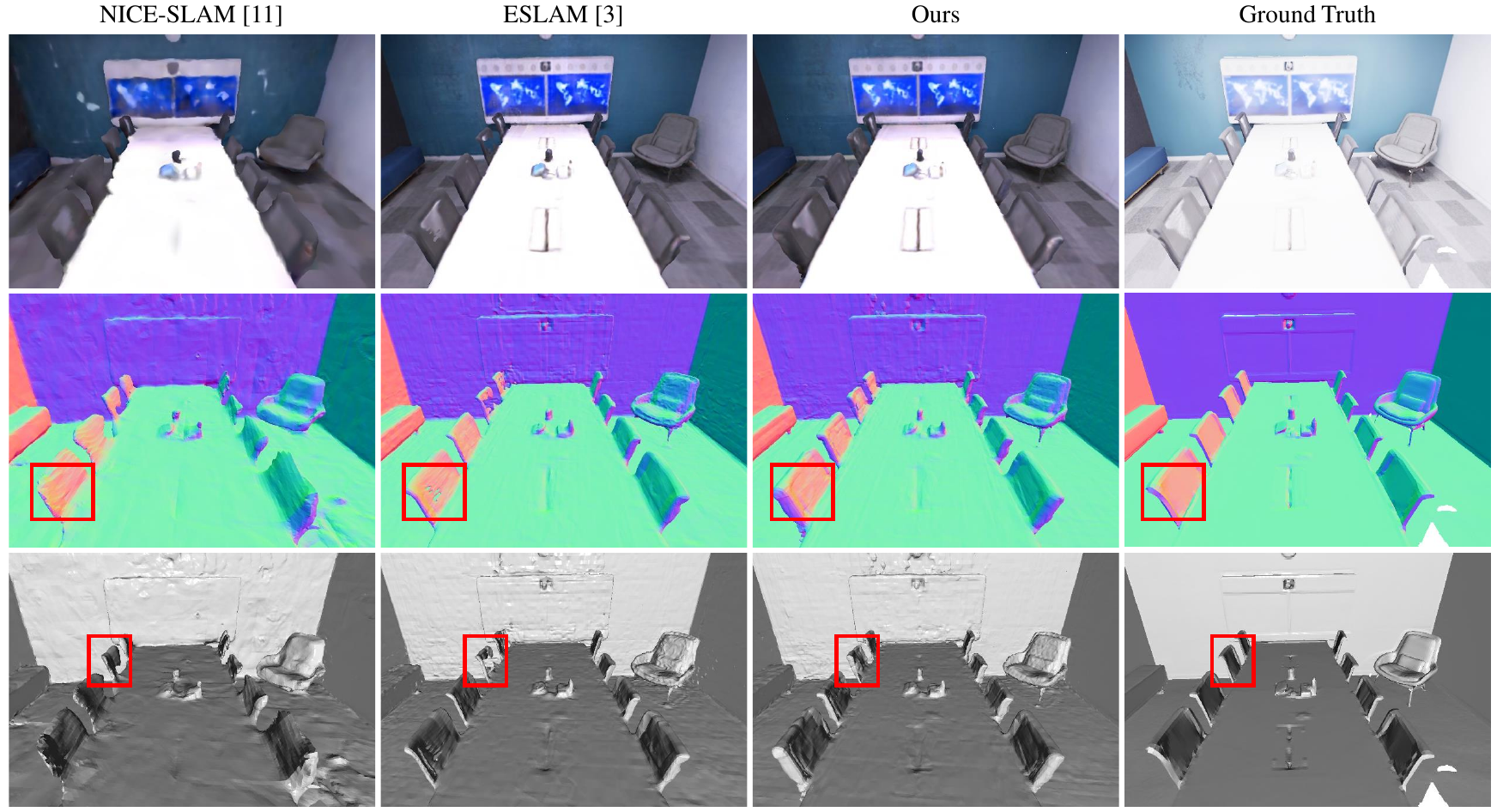}
  \caption{Qualitative comparison of our method with existing NeRF-based SLAM methods on Replica~\cite{straub2019replica} of office4 using different shading mode. As shown in \textcolor{red}{red colored box}, our method achieves complete reconstruction compared with other methods. }
  \label{fig:suppl_off4_detail}
  
\end{figure*}

\begin{figure*}
  \centering
  \includegraphics[width=\linewidth]{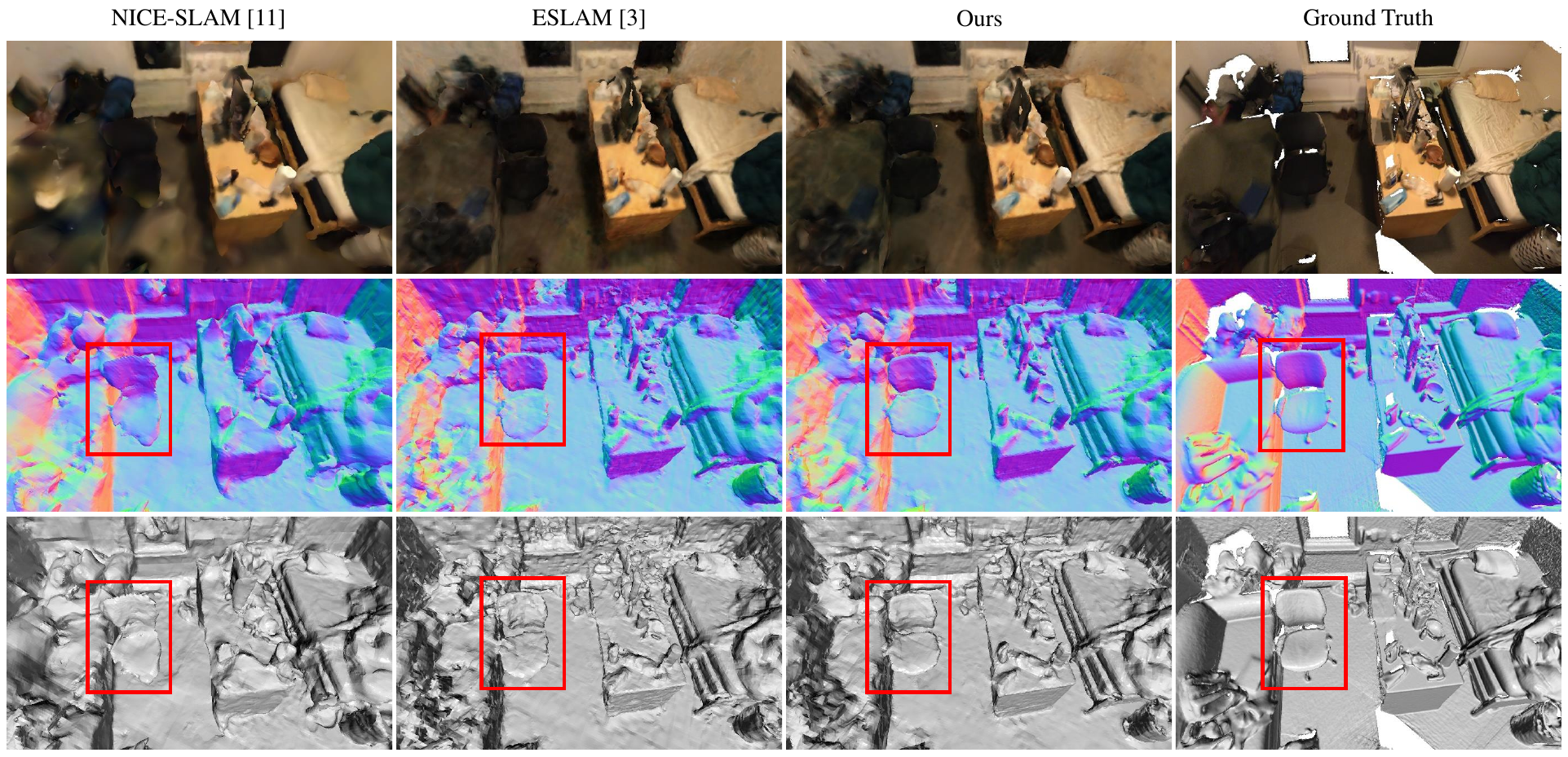}
  \caption{Qualitative comparison of our method with existing NeRF-based SLAM methods on ScanNet~\cite{dai2017scannet} of scene0207 using different shading mode. As shown in \textcolor{red}{red colored box}, our method achieves more accurate reconstruction compared with other methods. }
  \label{fig:suppl_scan207_detail}
\end{figure*}

Fig.~\ref{fig:suppl_off3_detail} and Fig.~\ref{fig:suppl_off4_detail} demonstrate detailed zoom-in views of Replica dataset~\cite{straub2019replica} and Fig.~\ref{fig:suppl_scan207_detail} shows results on ScanNet dataset~\cite{dai2017scannet}. While other methods fail to reconstruct details such as chair legs and chair back, our method is capable of complete scene reconstruction. This is attributed to the integration of different modality information, as it enables the complementarity of multi-modal information. 
Moreover, as shown in Fig.~\ref{fig:suppl_scan207_detail}, our method can achieve a smoother mesh reconstruction due to fully utilizing the advantages of each modality.
The above results indicate that our method is capable of retaining more details and achieving a more complete reconstruction compared to existing NeRF-based SLAM methods. Furthermore, our method can provide smoother, more coherent transitions for high-quality reconstruction.

\section{Video Demo}
\label{sec:Video Demo}
We present a video demo, demo.mp4, on room0 of the Replica dataset~\cite{straub2019replica}. In this video, we demonstrate the entire real-time mapping process, including semantic mapping and RGB mapping, as well as the tracking process, showing excellent performance of our method. From the video, we can view accurate semantic segmentation and RGB mapping results of the scene from top view. In addition, it can be observed that the ground truth trajectory and the estimated trajectory almost completely overlap, demonstrating high localization accuracy. We strongly recommend readers to view our video.

\vspace{2in}

\vspace{1in}

